\pdfoutput=1
\documentclass[5p,times,authoryear]{elsarticle}
\usepackage[T1]{fontenc}
\usepackage[utf8]{inputenc}
\usepackage{amsmath,amssymb,amsfonts,bm}
\usepackage{booktabs,array,tabularx,multirow}
\usepackage{graphicx,xcolor}
\usepackage{enumitem}
\usepackage{microtype}
\usepackage{xurl}
\usepackage[colorlinks=true,linkcolor=blue!40!black,citecolor=blue!40!black,urlcolor=blue!40!black]{hyperref}
\biboptions{authoryear,round}
\journal{Medical Image Analysis}
\graphicspath{{./}{fig1/}{fig_affiliation/}}

\setlist[itemize]{leftmargin=*,nosep}
\setlist[enumerate]{leftmargin=*}

\hypersetup{pdftitle={IC-GNN: Modeling Individualized Brain Community Organization for Disease Prediction},pdfauthor={Yang Zhang et al.},pdfsubject={Manuscript submitted to Medical Image Analysis}}

\begin{document}
\begin{frontmatter}
\title{IC-GNN: Modeling Individualized Brain Community Organization for Disease Prediction}
\author[yale-cbb]{Yang Zhang\fnref{equal}}
\author[yale-cbb]{Xiao Zhou\fnref{equal}}
\author[yale-cbb]{Jonathan Warrell\fnref{equal}}
\author[rutgers]{Avram Holmes}
\author[pratt]{Xuan Zhang}
\author[yale-cbb,yale-mbb,yale-cs]{Mark Gerstein\corref{cor}}
\ead{mark@gersteinlab.org}
\author{for the Alzheimer's Disease Neuroimaging Initiative\fnref{adni}}
\fntext[equal]{These authors contributed equally to this work.}
\cortext[cor]{Corresponding author.}
\fntext[adni]{Data used in preparation of this article were obtained from the Alzheimer's Disease Neuroimaging Initiative (ADNI) database (\href{https://adni.loni.usc.edu}{\nolinkurl{adni.loni.usc.edu}}). As such, the investigators within the ADNI contributed to the design and implementation of ADNI and/or provided data but did not participate in the analysis or writing of this report. A complete listing of ADNI investigators can be found at: \url{http://adni.loni.usc.edu/wp-content/uploads/how_to_apply/ADNI_Acknowledgement_List.pdf}}
\affiliation[yale-cbb]{organization={Program in Computational Biology and Bioinformatics, Yale University},city={New Haven},state={CT},postcode={06520},country={USA}}
\affiliation[rutgers]{organization={Department of Psychiatry, Robert Wood Johnson Medical School, and Brain Health Institute, Rutgers University},city={Piscataway},state={NJ},postcode={08854},country={USA}}
\affiliation[pratt]{organization={School of Design, Pratt Institute},city={New York},state={NY},postcode={11205},country={USA}}
\affiliation[yale-mbb]{organization={Department of Molecular Biophysics and Biochemistry, Yale University},city={New Haven},state={CT},postcode={06520},country={USA}}
\affiliation[yale-cs]{organization={Department of Computer Science, Yale University},city={New Haven},state={CT},postcode={06520},country={USA}}

\begin{abstract}
Brain networks exhibit modular organization, where community structures vary substantially across individuals and diseases. Understanding these individual differences is crucial for advancing both basic neuroscience research and precision medicine. However, widely used predefined group-level brain atlases often overlook this heterogeneity. While individualized functional parcellations capture subject-specific variability in community boundaries, they often require lengthy scan times to achieve fine-grained resolution and complicate cross-subject comparisons. To address this trade-off and investigate how individualized community organization captures disease-associated changes across individuals, we propose IC-GNN, an atlas-anchored model that forms distinct, individualized functional communities directly over atlas-defined brain regions. IC-GNN derives discrete, individualized multiscale communities from resting-state functional magnetic resonance imaging (fMRI) data, refines community-guided connections, and adapts shared regional weights to each subject's community context to enable both individualized modeling and group-level comparisons. Across more than 30,000 participants from ABIDE (autism spectrum disorder), ADNI (Alzheimer's disease), and the UK Biobank (depression), IC-GNN achieves the highest performance compared with 15 baseline methods. We further demonstrate that these individualized communities are reproducible within scans and across sessions, and carry predictive power on their own. Furthermore, IC-GNN identifies disease-related shifts in community organization that align with established literature, highlighting a promising approach for modeling individualized community structure for disease prediction and biomarker discovery.
\end{abstract}
\end{frontmatter}

\section{Introduction}
\label{sec:introduction}

Brain networks exhibit a modular organization, characterized by densely connected local communities with sparse connections between them~\citep{modular-brain}. Changes in community organization are often associated with neurological and psychiatric disorders, such as Alzheimer's disease (AD), major depressive disorder, and autism spectrum disorder (ASD)~\citep{ad-meta-review,Zhang2023,Lynch2024,Benkarim2021}. To study disease-associated changes in brain network organization, resting-state functional magnetic resonance imaging (fMRI) is widely used to indirectly measure neural activity through blood oxygenation fluctuations. Using fMRI, researchers have identified a broadly shared functional network organization across individuals~\citep{Yeo2011}. These findings have motivated the development of group-level brain atlases, which provide a common reference for brain network analysis~\citep{schaefer2018}.

\begin{figure}[t]
\centering
\includegraphics[width=\columnwidth]{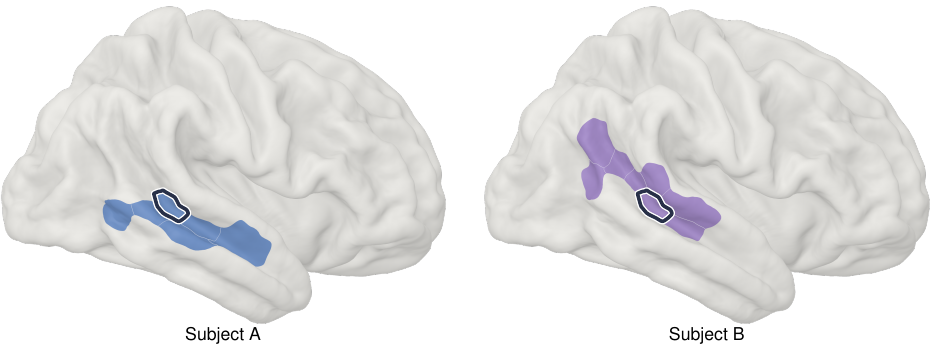}
\caption{An example of individual differences in functional community organization. The same outlined region of interest (RH\_TempPar\_5 in the Schaefer atlas, located in the right posterior superior temporal sulcus) belongs to communities comprising different sets of regions in Subject A (blue) and Subject B (purple).}
\label{fig:seed}
\end{figure}

However, different individuals may exhibit distinct community structures~\citep{Hermosillo2024}. Recent studies have increasingly emphasized shifting from group-level atlases to precise individualized brain mapping to advance both basic neuroscience research and precision medicine~\citep{Li2024}. Specifically, investigations in individualized functional parcellation have revealed reproducible individual differences in functional network topography, including the spatial distribution, boundaries, and regional membership of functional networks~\citep{Kong2019,Kong2021}. More importantly, variations in functional network organization have been associated with cognitive differences~\citep{Kong2019} and brain disorders, including depression~\citep{Lynch2024}. A direct way to capture such individual variability is to apply fine-grained individualized parcellations at the voxel level. However, estimating fine-grained individual boundaries can require extensive fMRI data, and differences in parcel definitions can complicate comparisons across subjects. To overcome this trade-off, we adopt an alternative strategy that preserves predefined brain regions while allowing them to form distinct functional communities in each individual. Even when brain regions are defined using a common atlas, how they group into functional communities may differ across individuals, as illustrated in Fig.~\ref{fig:seed}. This strategy characterizes individual differences in community membership at the scale of atlas-defined regions while retaining the same regional definitions across subjects.

Graph neural networks (GNNs) are widely used for brain disease prediction because of the inherent graph-like characteristics of brain networks~\citep{BrainGNN,BrainGB}. In the conventional pipeline, a brain atlas partitions preprocessed fMRI volumes into regions of interest (ROIs) that can be viewed as nodes. The pairwise statistical dependencies between the ROI-level blood-oxygen-level-dependent (BOLD) time series are then estimated to form a functional connectivity matrix that defines the edges of the graph. Existing methods have incorporated the modular organization of the brain through community-aware graph learning~\citep{BrainGNN,ComBrainTF,CAGT,BrainHGT}. However, incorporating community information into graph learning does not necessarily account for individual differences in regional community membership. Existing methods often rely on predefined community partitions or learn implicit community-aware representations to facilitate feature aggregation and information exchange. Subject-specific differences in how brain regions form communities receive less direct attention, and their role in disease prediction remains less well understood. To address this gap, we derive discrete, multiscale community memberships from each subject's functional connectivity and use them as an explicit structural prior for graph learning. This allows individual community organization to guide how brain connections are refined and how regional evidence contributes to disease prediction.

Building on this idea, we propose an individualized community-guided graph neural network (IC-GNN) that models individualized community organization within a common brain atlas. An affinity propagation-based hierarchical pooling module (AP-HPM) derives discrete community hierarchies from each subject's functional connectivity. These hierarchies retain the membership of each ROI across levels, allowing comparisons of community organization across scales and subjects. We also incorporate a probabilistic edge refinement module (PERM) that uses community membership to explicitly refine graph connections. Since the derived communities vary in number and composition across subjects, we design a community-guided readout that anchors disease prediction back to the original atlas-defined ROIs. Specifically, the readout adapts shared, region-specific readout weights to each subject's multiscale community context, allowing us to directly compare the predictive contribution of the same brain region across individuals. Our contributions can be summarized as follows:

\begin{enumerate}
\item We introduce a framework that captures individual variations in functional networks through the reconfiguration of fine-grained, atlas-defined ROIs into discrete multiscale communities. Unlike voxel-level individualized parcellations or implicit GNN cluster assignments, this formulation derives intrinsic, data-driven community structures while preserving strict region-to-region correspondence for cross-subject and clinical comparisons.

\item We develop IC-GNN, which integrates AP-HPM for multiscale community extraction, PERM for community-guided connection refinement, and a community-guided readout to adjust shared regional weights to each subject's community context. Across more than 30,000 participants from ABIDE (ASD), ADNI (AD), and the UK Biobank (depression), IC-GNN consistently achieves the highest diagnostic performance compared with 15 baseline methods.

\item We demonstrate that the derived individualized communities act as reproducible individual fingerprints within scans and across sessions and carry diagnostic information beyond demographic and acquisition covariates. Moreover, our framework reveals progressive, disease-associated disruptions in canonical network organization and shows that IC-GNN's additive regional evidence tracks these community reorganizations.
\end{enumerate}

\section{Related work}
\label{sec:related}

\subsection{Individualized functional brain organization}

The human brain exhibits a remarkable degree of individual variability in its functional organization. Conventional group-level functional brain atlases are often used to capture a common network architecture broadly shared across individuals~\citep{Yeo2011,schaefer2018}. Despite enabling meaningful comparisons across individuals, group-level atlases necessarily obscure individual differences in functional organization~\citep{Laumann2015,Gordon2017}. Subsequent studies have found that such differences are not merely transient noise, but reflect stable organization across sessions~\citep{Gratton2018}. They also contain additional information that can predict individual variation in cognition, personality, and emotion~\citep{Kong2019}. To characterize individual functional organization, two methodological paradigms have emerged. Individualized parcellation redefines cortical network boundaries for each subject to estimate fine-grained spatial topography~\citep{Wang2015,Li2017,Dickie2018,Kong2019,Kong2021}, which typically requires either extensive scan data~\citep{Laumann2015,Gratton2020} or group-level priors. The second paradigm keeps predefined atlas ROIs fixed and estimates how they assemble into individualized functional communities, which preserves direct region-to-region correspondence across a cohort~\citep{Liao2017,Salehi2018,Betzel2019,Akiki2019}. These ROI-level community assignments have been shown to exhibit reproducible patterns of individual variability~\citep{Liao2017}, capture individual differences~\citep{Salehi2018}, and reflect multiscale variability linked to distinct cognitive domains~\citep{Betzel2019}. Furthermore, studies across both paradigms indicate that perturbations in individualized network topography and community structure are closely associated with brain disorders, carrying disease-related signals across autism spectrum disorder~\citep{Keown2017,Dickie2018,Nunes2019,Benkarim2021}, Alzheimer's disease~\citep{Contreras2019}, and depression~\citep{Lynch2024,Pang2026}. Together, these findings motivate exploring how individualized functional organization can be leveraged in predictive models.

\subsection{Deep learning for functional brain networks}

Deep learning has been widely adopted to predict brain disorders from fMRI functional connectivity~\citep{BrainGB,brain-graph-survey,brainGNN-survey}. Prevalent approaches range from convolutional neural networks on connectivity matrices~\citep{BrainNetCNN} to graph neural networks and graph transformers~\citep{BrainGNN,BrainNetTransformer,ALTER}. Graph structure learning techniques are often employed to suppress noisy connections introduced by physiological artifacts and learn a refined graph structure. These methods typically learn graphs end-to-end from BOLD time series using task-driven~\citep{FBNetGen} or unsupervised objectives~\citep{BrainUSL}, or alternatively learn sparse edge masks over precomputed connectivity matrices~\citep{IBGNN}. However, the resulting connections are not guided by community structure. Recent models increasingly incorporate community structure into graph learning. Some methods rely on predefined community partitions shared across subjects to guide intra-community and inter-community feature interactions~\citep{ComBrainTF,CAGT}. Beyond fixed partitions, other approaches capture community structure implicitly~\citep{BQN} or learn community-aware representations through soft clustering, including ROI-aware graph convolutions~\citep{BrainGNN}, orthonormal cluster readouts~\citep{BrainNetTransformer}, and atlas-guided soft community assignments~\citep{BrainHGT}. More recently, BrainPICM explicitly estimates subject-specific soft community assignments to guide self-supervised masking and characterize atlas deviations for disease classification~\citep{BrainPICM}. While these studies demonstrate the value of incorporating community information into brain network learning, less attention has been paid to explicitly modeling individual differences in community organization and systematically investigating their predictive value and associations with brain disorders. In particular, interpretable brain network models report the evidence behind their predictions as salient ROIs~\citep{BrainGNN}, disorder-related connections~\citep{IBGNN,BrainIB}, or community-level attention~\citep{ComBrainTF,CAGT,BrainHGT}, but these analyses do not relate the evidence to each subject's community organization.

\section{Methods}
\label{sec:methods}

\begin{figure*}[t]
\centering
\includegraphics[width=\textwidth]{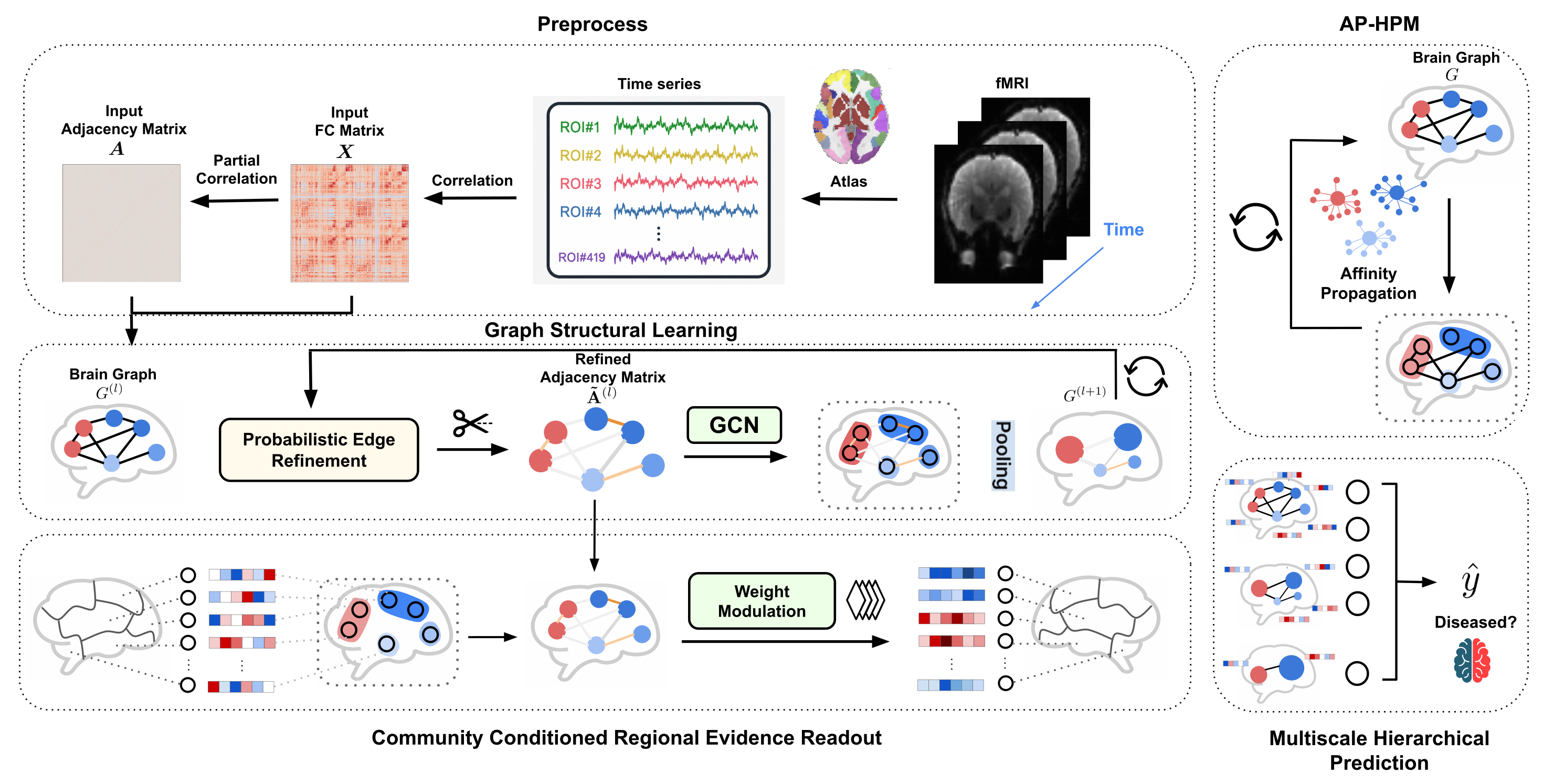}
\caption{The overall framework of IC-GNN.}
\label{fig:framework}
\end{figure*}

In this section, we introduce the overall architecture of IC-GNN. At a high level, AP-HPM first constructs discrete individualized hierarchical community assignments. At each hierarchical level, PERM refines graph connectivity using node embeddings and community membership, followed, except at the final level, by a graph convolutional operation that updates node embeddings. The updated embeddings are then pooled according to the precomputed community assignments. The learned multiscale connectivity matrices directly guide the subsequent community-guided readout. Specifically, the readout module uses each subject's refined topological structure to dynamically adapt shared ROI-specific prediction parameters at each level. Finally, the resulting multiscale ROI-level contributions are aggregated to produce the final prediction. The overall architecture is illustrated in Fig.~\ref{fig:framework}.

\subsection{Brain functional graph construction}

\label{sec:graph_construction}

Define an undirected, weighted graph as $G = (\bm{X}, \bm{A})$, where $\bm{X} \in \mathbb{R}^{n \times d}$ is the node feature matrix with $n$ nodes and $d$ dimensions, and $\bm{A} \in \mathbb{R}^{n \times n}$ is the weighted adjacency matrix. In our setting, $\bm{X}$ represents the input functional connectivity matrix, where $x_{ij}$ is the Fisher-transformed Pearson correlation coefficient of ROI $i$ with ROI $j$. The $i$-th row $x_i$, standardized to zero mean and unit variance, is the connectivity profile of ROI $i$.
Given a graph dataset $\mathcal{D}=\{(G_1, y_1), (G_2,y_2),\dots\}$, where $y_m\in \mathcal{Y} = \{0, 1\}$ is the disease label corresponding to graph $G_m\in \mathcal{G}$, our goal can be framed as a binary classification task to learn a model $f:\mathcal{G}\rightarrow\mathcal{Y}$ that maps each graph to a label.

\subsection{Subject-specific graph structural learning}
\label{sec:ap}

Individual functional organization determines which regions form communities and how these communities communicate. We represent this organization through a subject-specific community hierarchy, constructed by the affinity propagation-based hierarchical pooling module (AP-HPM). We index the hierarchy by level $l=0,\dots,L-1$ with $L=4$, where level $0$ is the ROI graph and levels $1$ to $L-1$ are community graphs. At each level, a probabilistic edge refinement module (PERM) adjusts the connections used for information exchange. Together, these components provide the community memberships and refined adjacency matrices that condition the community-guided readout described in the next subsection.

\subsubsection{Discrete brain community assignments}

At each level, we construct a similarity matrix $\bm{S}^{(l)}$ for each subject that measures the pairwise similarity between nodes based on their functional connectivity profiles. At the first level, the similarity matrix is obtained directly from the input functional connectivity matrix $\bm{X}$ as follows:
\begin{equation}
 S^{(0)}_{ij}=\exp\left(-\frac{\lVert x_{i}-x_{j}\rVert_1}{\operatorname{median}_{u\ne v}\lVert x_{u}-x_{v}\rVert_1}\right).
 \label{eq:affinity}
\end{equation}
The median normalizes the distances to a common scale for each subject. The obtained similarity matrix is then treated as the input to the affinity propagation (AP) algorithm~\citep{Frey2007} to obtain a discrete, subject-specific community assignment $\bm{C}^{(l)}\in\{0,1\}^{K^{(l)}\times K^{(l+1)}}$, where $K^{(l)}$ is the number of nodes at level $l$ and $K^{(0)}=n$. Because AP does not require a predefined number of communities, it allows community sizes and membership to vary across subjects.

At deeper levels, the similarity $\bm{S}^{(l)}$ between communities is defined as the average similarity between their members. The similarity matrix is pooled hierarchically from the previous level. Let $\bm D_{C}^{(l)}$ denote the diagonal matrix containing the number of level-$l$ nodes assigned to each community at level $l+1$. The pooled similarity matrix can be computed as:
\begin{equation}
 \bm S^{(l+1)}=(\bm D_{C}^{(l)})^{-1}(\bm C^{(l)})^\top \bm S^{(l)} \bm C^{(l)}(\bm D_{C}^{(l)})^{-1}.
 \label{eq:affinitypool}
\end{equation}
Each entry of the pooled similarity matrix represents the average connectivity-profile similarity between the members of two communities. AP is then applied recursively to the pooled similarity matrices to construct three hierarchical community levels.

\subsubsection{Community-guided edge refinement}

Physiological noise and preprocessing artifacts can introduce unreliable correlations into fMRI-derived graphs. Their influence can propagate through message passing and obscure disease-related information. To reduce the influence of the noisy connections and improve interpretability, we propose a probabilistic edge refinement module (PERM) that learns to attenuate potentially uninformative connections, guided by node embeddings and subject-specific community membership.

At each level, PERM combines connectivity-profile similarity, community membership, and current node embeddings to estimate a soft edge-keeping probability matrix $\bm{P}^{(l)}$. For levels $l=0,1,2$,
\begin{equation}
  P^{(l)}_{ij}=\sigma\big(w^{(l)}_a S^{(l)}_{ij}
 +w^{(l)}_b Q^{(l)}_{ij} + w^{(l)}_c \cos(H^{(l)}_{i}, H^{(l)}_{j})+ w^{(l)}_d\big),
 \label{eq:gate}
\end{equation}
where $\sigma(\cdot)$ is the sigmoid function, $ Q^{(l)}_{ij} \in \{-1,1\}$ indicates whether nodes $i$ and $j$ share a parent community, and $\cos(H^{(l)}_{i}, H^{(l)}_{j})$ measures the cosine similarity between their current embeddings. PERM is a self-explainable module. The learnable scalars $w^{(l)}_a, w^{(l)}_b, w^{(l)}_c, w^{(l)}_d$ make the influence of each factor on edge refinement explicit and inspectable at each hierarchical level. A higher $P^{(l)}_{ij}$ favors retaining more of the original edge weight. At the final level ($l=3$), PERM only uses connectivity-profile and embedding similarity because there is no parent community.

During training, PERM adopts a relaxed Bernoulli sampling scheme~\citep{D-CoRP} to allow the final refined connections to vary across training iterations, which reduces the model's reliance on a fixed set of edges for decision-making. We sample a relaxed mask by smoothly thresholding the probability $P^{(l)}_{ij}$ at each level against a uniform random variable $\epsilon_{ij}\sim\mathcal U(0,1)$:
\begin{equation}
 M^{(l)}_{ij}
 =
 \frac{1+\tanh\left(
 (P^{(l)}_{ij}-\epsilon_{ij})/\tau
 \right)}{2}.
 \label{eq:mask}
\end{equation}
The temperature $\tau$ controls the smoothness of the mask. As $\tau$ approaches $0$, the mask becomes equivalent to a binary Bernoulli mask. At a larger temperature, PERM provides continuous attenuation. The mask is rescaled to the range $[0.1,1]$, so that every edge retains at least 10\% of its weight. The refined adjacency is obtained by element-wise multiplication of the original adjacency matrix with the stochastic mask:
\begin{equation}
 \tilde A^{(l)}_{ij}
 =
 A^{(l)}_{ij}M^{(l)}_{ij}.
 \label{eq:refinedadj}
\end{equation}
During evaluation, we use the probability $P^{(l)}_{ij}$ as a deterministic mask.

\subsubsection{Graph convolutional layer}

We adopt the graph convolutional network (GCN)~\citep{GCN} as the message propagation operator in our model. Graph convolution uses the refined adjacency matrix $\tilde{\bm{A}}^{(l)}$ to update the node embeddings at levels $l=0,1,2$, with the first level starting from the input features $\bm H^{(0)}=\bm X$. Given the node embeddings $\bm{H}^{(l)}$ at level $l$, the updated embeddings $\tilde{\bm H}^{(l)}$ are computed through a sequence of graph convolution, normalization, and nonlinear Gaussian error linear unit (GELU) activation~\citep{GELU}:
\begin{equation}
 \tilde{\bm H}^{(l)}
 =
 \mathrm{GELU}\!\left(
 \mathrm{Norm}\!\left(
 \mathrm{GCN}
 \big(\bm H^{(l)},\tilde{\bm A}^{(l)}\big)
 \right)\right).
 \label{mp}
\end{equation}
For each graph convolution, we use standard adjacency normalization with unit self-loops added to the refined adjacency matrix $\tilde{\bm A}^{(l)}$, which preserves each node's own information when neighboring connections are attenuated.

\subsubsection{Hierarchical graph pooling}

Following graph convolution, the graph is coarsened according to the precomputed subject-specific community assignments $\bm C^{(l)}$. The coarsened adjacency matrix is obtained by aggregating connections between communities:
\begin{equation}
\bm A^{(l+1)} =
\operatorname{offdiag}\left(
(\bm{C}^{(l)})^\top \bm A^{(l)} \bm{C}^{(l)}
\right).
\end{equation}
Here, $\operatorname{offdiag}$ sets the diagonal entries to zero. The intra-community connectivity is summarized by the mean edge weight and edge density as node-level topological features in $\bm T^{(l+1)}$. This allows subsequent graph learning to capture interactions between communities while retaining their internal connectivity characteristics as node-level features.

The node features of the coarsened graph are obtained by averaging the updated embeddings of member nodes and concatenating the topological features:
\begin{equation}
 \bm H^{(l+1)}
 =
 \left[
 (\bm D_{C}^{(l)})^{-1}
 (\bm C^{(l)})^\top\tilde{\bm H}^{(l)}
 \;\Vert\;
 \bm T^{(l+1)}
 \right].
 \label{eq:hiddenpool}
\end{equation}
The resulting features are used for edge refinement and graph convolution at the next level, and only for edge refinement at the final level. Repeating this process across three pooling stages produces a hierarchy of progressively coarser graph representations.

\subsection{Community-guided regional evidence readout}
\label{sec:readout}

The refined multiscale connectivity matrices from the previous stage capture each subject's hierarchical structural context. However, because the derived communities vary in number and composition across individuals, we introduce a community-guided readout that anchors prediction back to the atlas-defined ROIs, preserving strict region-to-region correspondence and fine-grained interpretability. Specifically, the readout maintains cohort-shared, ROI-specific prediction parameters and dynamically modulates them at each hierarchical level using the subject's community organization and refined connectivity. This allows the same anatomical ROI to exert distinct predictive contributions across subjects, which are ultimately aggregated within each scale and combined across the hierarchy for the final prediction.

\subsubsection{Community-guided readout prototype propagation}

To achieve this, we first project the level-0 graph-convolved node representations $\tilde{\bm{H}}^{(0)}$ (Eq.~\ref{mp}) into a $d_z$-dimensional readout space via a linear layer: $\bm{Z} = \mathrm{MLP}_z(\tilde{\bm{H}}^{(0)}) \in \mathbb{R}^{n \times d_z}$, where the $i$-th row $Z_i \in \mathbb{R}^{d_z}$ denotes the projected feature of ROI $i$. We then introduce a cohort-shared ROI weight matrix $\bm{\Theta} \in \mathbb{R}^{n \times d_z}$, whose $i$-th row $\Theta_i$ provides the baseline predictive weights for $Z_i$. To adapt these shared weights to each subject's community context, we define the mapping matrix
\begin{equation}
  \bm U^{(l)}= \bm C^{(0)}\bm C^{(1)}\cdots \bm C^{(l-1)},
\end{equation}
where $\bm U^{(0)}=\bm I$, and $U_{ik}^{(l)}\in\{0,1\}$ indicates whether ROI $i$ belongs to community $k$ at level $l$. The community prototypes $\bar{\bm{\Theta}}^{(l)}$ are obtained by averaging the ROI weights within each community. As each subject differs in community membership, these pooled prototypes also vary across subjects. For each subject at level $l$,
\begin{equation}
 \bar{\bm{\Theta}}^{(l)}=(\bm D_U^{(l)})^{-1}(\bm U^{(l)})^\top \bm \Theta,
 \label{eq:prototype}
\end{equation}
where $\bm D_U^{(l)}$ is the diagonal matrix containing the number of ROIs in each community at level $l$.

The refined adjacency matrix $\tilde{\bm A}^{(l)}$ is used to modulate information exchange over the subject-specific community graph. The pooled community prototypes $\bar{\bm\Theta}^{(l)}$ first exchange information over $\tilde{\bm A}^{(l)}$ and are then broadcast to their member ROIs through the mapping $\bm U^{(l)}$:
\begin{equation}
 \tilde{\bm{\Theta}}^{(l)}=\bm U^{(l)} \hat{\bm A}^{(l)} \bar{\bm\Theta}^{(l)},
 \label{eq:context}
\end{equation}
where $\hat{\bm A}^{(l)}$ is $\bm I+\tilde{\bm A}^{(l)}$ followed by a row normalization. The unit self-loop retains each prototype's own contribution. The mapping $\bm U^{(l)}$ ensures that the exchanged information is returned to the original ROI level. The resulting $\tilde{\bm{\Theta}}^{(l)}$ therefore contains ROI-level prototypes that have been modulated by the subject's community context.

\subsubsection{Subject-specific weight modulation}
Having obtained the context-modulated ROI prototypes, we then adapt the original ROI weights $\bm \Theta$ to produce subject-specific weights $\Omega_{i}^{(l)}$ for each ROI $i$ at level $l$:
\begin{equation}
 \Omega_{i}^{(l)}=\Theta_i\odot[1+\mathrm{MLP}^{(l)}_{\gamma}(\tilde{\Theta}_{i}^{(l)})]+\mathrm{MLP}^{(l)}_{\beta}(\tilde{\Theta}_{i}^{(l)}).\label{eq:omega}
\end{equation}
Specifically, the final ROI-specific weights are obtained by modulating the original shared weights $\Theta_i$ with a multiplicative term and an additive term~\citep{FiLM}, both derived from the context-modulated prototypes $\tilde{\Theta}_{i}^{(l)}$ through separate linear layers. $\Omega_{i}^{(l)}$ serves as the final ROI-specific weight for the readout of the ROI embedding $Z_{i}$ at each level $l$. In this manner, each subject has its own ROI-specific weights that are informed by the subject's community context.

\subsubsection{Multiscale hierarchical prediction}

At every level, the process repeats to produce a set of ROI-specific weights $\Omega_{i}^{(l)}$ for each ROI. For every subject, the logit at each level is computed as the dot product of the subject-specific weights and the corresponding ROI embeddings. The final logit is obtained by summing the contributions from all ROIs across levels and is then passed through a sigmoid function for binary classification:
\begin{equation}
\hat y = \sigma\left(w_f + \sum_{l=0}^{L-1}w^{(l)}_e \sum_{i=1}^{n}\langle\Omega_{i}^{(l)},Z_{i}\rangle\right),
 \label{eq:logit}
\end{equation}
where $w_f$ is a learnable bias. The weights $w_e^{(l)}$ at each level are learned during training and softmax-normalized, allowing the model to adaptively learn the contributions from different levels. We refer to $c_i=\sum_{l=0}^{L-1}w^{(l)}_e\langle\Omega_{i}^{(l)},Z_{i}\rangle$ as the regional evidence of ROI $i$, so that the logit equals $w_f+\sum_{i=1}^{n}c_i$. Essentially, the model allows ROIs with the same features but different community contexts to contribute differently to the final prediction, which is crucial for capturing individual variability in functional brain organization across levels.

\subsection{Loss function}
\label{sec:loss}

We train IC-GNN using a standard class-weighted classification loss, $\mathcal{L}_{\mathrm{cls}}$, together with a regularization term on the PERM edge-keeping probabilities. The classification loss is defined as the binary cross-entropy between the predicted probability $\hat y$ and the true label $y$:
\begin{equation}
 \mathcal L_{\mathrm{cls}}
 =-\frac{1}{|\mathcal B|}\sum_{G\in\mathcal B}
 \left[
 \alpha y\log\hat y
 +(1-y)\log(1-\hat y)
 \right],
 \label{eq:classificationloss}
\end{equation}
where $\mathcal B$ is a training batch and $\alpha$ is the ratio of controls to disease cases in the training subset.

To control edge retention across levels, we regularize the edge-keeping probabilities toward a Bernoulli prior with keeping probability $r$:
\begin{equation}
 \mathcal L_{\rm edge}
 = \frac{1}{L} \sum_{l=0}^{L-1} \mathbb{E}_{(i,j) \sim \mathcal E^{(l)}} \left[ D_{\rm{KL}}^{\rm{Bern}} \left( P_{ij}^{(l)} \;\Vert\; r \right) \right],
\end{equation}
where $\mathcal E^{(l)}$ is the set of existing edges at level $l$ and $D_{\rm{KL}}^{\rm{Bern}}$ is the Kullback--Leibler divergence between two Bernoulli distributions. Combining this structural regularization with the cross-entropy classification loss $\mathcal{L}_{\mathrm{cls}}$, we formulate the overall training objective as:
\begin{equation}
 \mathcal L = \mathcal L_{\rm cls} + \lambda \mathcal L_{\rm edge},
\end{equation}
where $\lambda > 0$ is a hyperparameter balancing diagnostic performance and edge sparsity regularization.

\section{Experiments}
\label{sec:experiments}

\subsection{Datasets and preprocessing}

We evaluate IC-GNN on three public datasets: the Autism Brain Imaging Data Exchange (ABIDE)~\citep{ABIDE2014} for ASD, the Alzheimer's Disease Neuroimaging Initiative (ADNI)~\citep{ADNI2008} for AD, and UK Biobank~\citep{ukbiobank} for depression. All scans are parcellated into the same 419 regions, including 400 cortical regions from the Schaefer atlas (17-network version)~\citep{schaefer2018} and 19 noncortical regions following~\cite{ukbiobank_pipeline}. For ABIDE and ADNI, regional time series were extracted by averaging the voxel signals within each region, and the functional connectivity matrix consists of the Pearson correlations between all pairs of regional time series. For UK Biobank, we used precomputed Pearson functional connectivity matrices with the same parcellation. The node features of a region are its row of the Fisher-transformed functional connectivity matrix, standardized to zero mean and unit variance. The weighted adjacency matrices are constructed from ridge-regularized partial correlations, in which each region retains its $30\%$ strongest positive partial correlations. The union of these selections forms an undirected graph.

\paragraph{ABIDE}
The cohort included 884 participants from 20 acquisition sites. In total, there were 427 individuals with ASD and 457 healthy controls. Preprocessed resting-state fMRI data were obtained from the ABIDE Preprocessed Initiative using the C-PAC \texttt{nofilt\_noglobal} outputs~\citep{ABIDEPreprocessed}. The preprocessing procedures included slice-timing and motion correction, spatial normalization to MNI152 space at 3-mm resolution, and nuisance regression using 24 motion parameters, five CompCor components, and linear and quadratic trends. No band-pass filtering or global signal regression was applied. We excluded scans in which any region lay outside the functional field of view and kept only one of two participant identifiers that pointed to the same image.

\paragraph{ADNI}
Data used in the preparation of this article were obtained from the ADNI database (\href{https://adni.loni.usc.edu}{\nolinkurl{adni.loni.usc.edu}}). ADNI was launched in 2003 as a public-private partnership, led by Principal Investigator Michael W. Weiner, MD. The primary classification cohort comprised 1,114 participants from 64 sites, including 197 AD patients and 917 cognitively normal (CN) controls. An additional 618 participants with mild cognitive impairment (MCI) were not used for classification. Instead, they were included in the community analyses (Sections~\ref{sec:subject-communities} and~\ref{sec:disease-communities}). Structural and resting-state functional MRI data were preprocessed using fMRIPrep 23.2.1~\citep{fmriprep}, with functional images registered to MNI space at 2-mm resolution. Regional time series were further denoised using Nilearn with motion, mean white-matter and cerebrospinal-fluid signals, and cosine drift regressors, without additional band-pass filtering or global signal regression. For every participant, the scan at the earliest session with a matched diagnosis was used for classification.

\paragraph{UK Biobank}
The cohort comprised 28,602 participants, including 5,772 individuals reporting a lifetime professional diagnosis of depression and 22,830 controls reporting none of the listed mental-health conditions. We used precomputed functional connectivity matrices generated using the pipeline described by~\cite{ukbiobank_pipeline}, which included motion correction, high-pass filtering, ICA-FIX denoising, and global signal regression. Node features and edges were then computed as for the other two datasets.

\subsection{Training and evaluation protocol}

We validated our results across multiple random seeds. For ABIDE and ADNI, we performed five-fold site-grouped cross-validation, with acquisition sites separated across the training, validation, and test subsets. For UK Biobank, we fixed a held-out test set of 4,289 participants, including 866 depression cases. Within the development set of the remaining 24,313 participants, five stratified training and validation folds were formed, and each resulting model was evaluated on the same test set.

IC-GNN was optimized using the Adam optimizer~\citep{adam} on all datasets. The PERM parameters used learning rates of $9\times10^{-4}$, $9\times10^{-3}$, and $4.5\times10^{-3}$ for ABIDE, ADNI, and UK Biobank, respectively. PERM and the community-guided readout parameters used zero weight decay. In all datasets, the preference of AP at the first, second, and third levels is set to the $0.90$, $0.93$, and $0.90$ quantiles, respectively, of the off-diagonal entries of that level's similarity matrix. The hidden dimensions of learned embeddings are set to 256, 128, and 64, and the ROI readout dimension $d_z$ is 8. Training ran for up to 300 epochs. ABIDE and ADNI used a five-epoch linear warm-up followed by cosine decay, while UK Biobank used a ten-epoch warm-up followed by cosine annealing with restarts every ten epochs. Early stopping was based on the validation area under the receiver operating characteristic curve (AUROC). The trained models were selected using exponential moving averages of model parameters, with decay rates of 0.99 for ABIDE and ADNI and 0.9985 for UK Biobank. For threshold-dependent metrics, ABIDE and ADNI used validation-selected Youden thresholds. UK Biobank used a common operating point, with each model's threshold selected on its validation scores so that 37.5\% of its validation participants are classified as positive. The same threshold rules were applied to every method (Supplementary Section~\ref{sup:bench-protocol}).

\section{Results}
\label{sec:results}

\subsection{Comparisons with the state-of-the-art methods}

We compare IC-GNN with various types of baselines: (1) conventional machine learning methods, including Support Vector Machine (SVM) and Random Forest; (2) general-purpose GNN backbones, including GCN~\citep{GCN}, GAT~\citep{GAT}, GIN~\citep{GIN}, and GraphSAGE~\citep{GraphSAGE}; and (3) deep learning methods for brain networks, including BrainNetCNN~\citep{BrainNetCNN}, BrainGNN~\citep{BrainGNN}, BrainNetTF~\citep{BrainNetTransformer}, ASD-DiagNet~\citep{ASDDiagNet}, BQN~\citep{BQN}, Com-BrainTF~\citep{ComBrainTF}, BrainHGT~\citep{BrainHGT}, ALTER~\citep{ALTER}, and CAGT~\citep{CAGT}.

We report the accuracy and AUROC in Table~\ref{tab:accuracy_three_datasets}. Among all the models we compared, including the previous state-of-the-art methods, IC-GNN has the highest mean AUROC and accuracy on all three datasets. In particular, IC-GNN obtains an AUROC of 75.41\% and an accuracy of 69.40\% on ABIDE. On ADNI, IC-GNN achieves an AUROC of 86.46\% and an accuracy of 80.52\%. We also test our model on UK Biobank with a much larger sample size than ADNI and ABIDE, where IC-GNN maintains its leading performance and achieves an AUROC of 60.62\% and an accuracy of 62.51\%.

We also observe that the conventional SVM has a higher mean AUROC than all general-purpose GNNs. This is consistent with a recent study~\citep{Han2026}, which found that the message-aggregation mechanism of graph deep learning models did not help the predictive performance for functional connectomes. One possible explanation is that repeated graph aggregation and pooling may weaken subtle disease-related signals at the ROI level. IC-GNN is designed to avoid this loss by using the subject-specific community structures to adapt ROI-specific prediction weights, while retaining the original ROI-level evidence for classification. We also report additional metrics, including balanced accuracy, specificity, and precision, recall, and F1 for the disease-positive class, in Supplementary Section~\ref{sup:bench}. IC-GNN achieves the highest mean F1 score on both ABIDE (63.55\%) and ADNI (58.04\%) and the second highest on UK Biobank.

\begin{table*}[t]
\centering
\small
\setlength{\tabcolsep}{3.5pt}
\renewcommand{\arraystretch}{1.10}
\caption{Classification performance on ABIDE (ASD vs.\ controls), ADNI (AD vs.\ CN), and UK Biobank (depression vs.\ controls). Values are percentages (mean $\pm$ standard deviation) over the five folds. The best result in each column is in bold and the second best is underlined.}
\label{tab:accuracy_three_datasets}
\begin{tabular}{@{}llcccccc@{}}
\toprule
\multirow{2}{*}{Category} & \multirow{2}{*}{Model} & \multicolumn{2}{c}{ABIDE} & \multicolumn{2}{c}{ADNI} & \multicolumn{2}{c}{UK Biobank} \\
\cmidrule(lr){3-4}\cmidrule(lr){5-6}\cmidrule(lr){7-8}
& & ACC & AUROC & ACC & AUROC & ACC & AUROC \\
\midrule
\multirow{2}{*}{ML methods} & SVM & $63.49 \pm 2.42$ & $71.42 \pm 5.34$ & $71.22 \pm 4.38$ & $82.54 \pm 2.21$ & $61.70 \pm 0.53$ & $59.06 \pm 0.28$ \\
 & Random Forest & $61.73 \pm 3.14$ & $65.85 \pm 4.96$ & $65.64 \pm 6.96$ & $70.25 \pm 4.40$ & $57.01 \pm 1.11$ & $51.69 \pm 0.28$ \\
\midrule
\multirow{4}{*}{General GNNs} & GCN~{\footnotesize\citep{GCN}} & $58.60 \pm 3.76$ & $61.46 \pm 5.61$ & $69.52 \pm 8.51$ & $74.89 \pm 2.90$ & $60.35 \pm 0.40$ & $57.04 \pm 0.46$ \\
 & GAT~{\footnotesize\citep{GAT}} & $58.52 \pm 3.43$ & $62.97 \pm 5.98$ & $73.02 \pm 7.76$ & $77.59 \pm 3.33$ & $60.83 \pm 0.20$ & $57.12 \pm 0.11$ \\
 & GIN~{\footnotesize\citep{GIN}} & $52.34 \pm 5.98$ & $56.54 \pm 4.34$ & $66.36 \pm 5.17$ & $71.98 \pm 2.99$ & $59.99 \pm 0.54$ & $55.52 \pm 0.62$ \\
 & GraphSAGE~{\footnotesize\citep{GraphSAGE}} & $59.51 \pm 5.01$ & $62.53 \pm 5.36$ & $68.93 \pm 7.32$ & $76.54 \pm 1.41$ & $60.41 \pm 0.54$ & $57.60 \pm 0.31$ \\
\midrule
\multirow{9}{*}{Brain models} & BrainNetTF~{\footnotesize\citep{BrainNetTransformer}} & $66.41 \pm 4.70$ & $\underline{73.84} \pm 6.19$ & $\underline{78.56} \pm 6.06$ & $83.98 \pm 1.81$ & $59.36 \pm 0.82$ & $54.13 \pm 1.75$ \\
 & BrainNetCNN~{\footnotesize\citep{BrainNetCNN}} & $65.74 \pm 3.83$ & $70.91 \pm 5.15$ & $71.69 \pm 7.04$ & $81.20 \pm 4.59$ & $60.86 \pm 0.77$ & $58.04 \pm 0.64$ \\
 & BrainGNN~{\footnotesize\citep{BrainGNN}} & $55.30 \pm 3.49$ & $59.18 \pm 5.24$ & $65.24 \pm 7.02$ & $75.13 \pm 1.51$ & $59.96 \pm 0.69$ & $56.98 \pm 0.33$ \\
 & ASD-DiagNet~{\footnotesize\citep{ASDDiagNet}} & $\underline{66.98} \pm 4.77$ & $71.74 \pm 7.74$ & $77.29 \pm 5.30$ & $\underline{84.78} \pm 3.59$ & $60.80 \pm 0.58$ & $58.07 \pm 0.47$ \\
 & BQN~{\footnotesize\citep{BQN}} & $63.55 \pm 4.63$ & $68.24 \pm 5.48$ & $71.54 \pm 5.81$ & $79.50 \pm 3.38$ & $61.58 \pm 0.75$ & $57.68 \pm 0.77$ \\
 & Com-BrainTF~{\footnotesize\citep{ComBrainTF}} & $66.03 \pm 4.81$ & $70.38 \pm 8.04$ & $74.88 \pm 6.53$ & $82.29 \pm 5.85$ & $\underline{62.00} \pm 0.50$ & $\underline{60.35} \pm 0.56$ \\
 & BrainHGT~{\footnotesize\citep{BrainHGT}} & $61.98 \pm 3.19$ & $69.48 \pm 8.00$ & $75.14 \pm 6.41$ & $84.51 \pm 2.84$ & $61.39 \pm 0.42$ & $59.34 \pm 0.78$ \\
 & ALTER~{\footnotesize\citep{ALTER}} & $66.20 \pm 2.14$ & $70.73 \pm 5.19$ & $74.28 \pm 6.91$ & $84.10 \pm 2.72$ & $61.94 \pm 0.69$ & $60.10 \pm 0.43$ \\
 & CAGT~{\footnotesize\citep{CAGT}} & $62.61 \pm 2.30$ & $65.91 \pm 5.94$ & $76.28 \pm 5.07$ & $80.90 \pm 2.89$ & $61.89 \pm 0.94$ & $59.16 \pm 0.66$ \\
\midrule
Our model & \textbf{IC-GNN} & $\mathbf{69.40 \pm 4.23}$ & $\mathbf{75.41 \pm 5.73}$ & $\mathbf{80.52 \pm 5.16}$ & $\mathbf{86.46 \pm 3.03}$ & $\mathbf{62.51 \pm 0.67}$ & $\mathbf{60.62 \pm 0.45}$ \\
\bottomrule
\end{tabular}
\end{table*}

\subsection{Ablation studies}

In this section, we conduct ablation studies to evaluate the contributions of different components in IC-GNN. We focus on the ABIDE dataset for these experiments. We investigate two main components: (1) the community hierarchy and (2) the edge refinement in PERM. The results of the ablation studies are summarized in Table~\ref{tab:ablation_abide}, and each ablation setting is described in Supplementary Section~\ref{sup:abl}.

First, to test whether the community structure helps, we remove the hierarchical structure that aims to model interactions from the ROI level to the community levels. We observe that the performance drops substantially when the hierarchy is removed. We also compare the performance of IC-GNN using subject-specific communities generated by AP against two baselines: a consensus hierarchy built from the training participants and a size-matched random hierarchy. The results show that the subject-specific communities perform the best. These results suggest that subject-specific community assignments are helpful for model learning.

Second, we evaluate the edge refinement in PERM. We compare the performance of the model using learned edge refinement with models using fixed open edges and uniform attenuation at the learned keep rate. The results show a performance drop when the learned edge refinement is removed, particularly in accuracy. This suggests that learned edge refinement contributes to the model's classification performance.

\begin{table*}[t]
\centering
\small
\renewcommand{\arraystretch}{1.12}
\caption{Ablation results of IC-GNN on ABIDE. Values are percentages (mean $\pm$ standard deviation) over the five test folds. The best results are in bold.}
\label{tab:ablation_abide}
\begin{tabular}{@{}lcllcc@{}}
\toprule
Comparison & Hierarchy & Community & Edge & AUROC & ACC \\
\midrule
Full model
& Yes & Individual & Learned
& $\bm{75.41 \pm 5.73}$
& $\bm{69.40 \pm 4.23}$ \\
\midrule
Hierarchy removal
& No & -- & Learned
& $71.68 \pm 6.29$ & $65.32 \pm 5.89$ \\
\midrule
Community control
& Yes & Random & Learned
& $73.79 \pm 3.67$ & $67.49 \pm 2.91$ \\
& Yes & Consensus & Learned
& $74.65 \pm 4.49$ & $68.17 \pm 2.58$ \\
\midrule
Edge control
& Yes & Individual & Open
& $74.56 \pm 4.90$ & $68.17 \pm 3.54$ \\
& Yes & Individual & Uniform
& $75.16 \pm 4.23$ & $67.14 \pm 2.85$ \\
\midrule
Joint control
& No & -- & Open
& $70.72 \pm 6.91$ & $65.66 \pm 5.47$ \\
\bottomrule
\end{tabular}
\end{table*}

\subsection{Reproducibility and atlas alignment of individualized  communities}
\label{sec:subject-communities}
In this section, we examine whether individualized communities capture reproducible individual differences rather than random scan noise. We first show that these communities are reproducible both within a single scan and across scanning sessions, preserving sufficient individual specificity to identify participants. We then show that they align with a canonical group-level atlas, and that this alignment strengthens further when communities are aggregated across participants.

For each ABIDE participant, we split the time series into two contiguous halves and generate individualized communities using AP from each half following the same procedure. We measure the similarity between two sets of communities by the Jaccard index of their co-assigned region pairs at the first (finest) community level. For each participant, we then search the second-half communities of all participants from the same site for the one most similar to the participant's first-half communities. The search returns the participant's own communities for 97.2\% of the participants, compared with 3.8\% when the first-half communities are replaced by random communities of the same sizes. This indicates that the community assignments are reproducible within a scan and specific enough to serve as an individual fingerprint. On ADNI, 980 participants have a second usable session, acquired a median of 14 months apart from the analyzed scan. We generate communities from both sessions and repeat the search within sites. The search returns the participant's own session for 71.9\% of the participants, compared with 6.3\% for random communities. This indicates that the community assignments contain individualized information across sessions, which is consistent with the findings in ABIDE.

We then examine how closely these individualized communities follow the functional networks of a group-level atlas. At the first community level, we compare the communities of each participant, using normalized mutual information (NMI), with a reference that assigns the cortical regions to the 17 networks of the Yeo atlas~\citep{Yeo2011} and the noncortical regions to one additional label. The mean NMI is 0.571 for ABIDE, 0.535 for ADNI (using all 1,732 participants), and 0.557 for UK Biobank, compared with about 0.35 for random communities of the same sizes. When the co-assignments of all participants in a cohort are averaged and clustered into a consensus partition, the NMI increases to 0.700, 0.694, and 0.726, respectively, which is higher than that of any single participant (Supplementary Section~\ref{sup:comm}). These results indicate that individualized communities capture both shared group-level organization and reproducible individual differences. This combination is also consistent with our readout design, which adapts shared ROI-specific weights using each subject's community context.

\subsection{Diagnostic information in individualized communities}
\label{sec:affiliation-prediction}
In this section, we investigate whether individualized community assignments contain diagnostic information beyond demographic and acquisition covariates.

To compare community assignments across participants, we represent each participant's communities using a common affiliation profile based on the Yeo-17 atlas and an additional noncortical system. For each ROI at each of the three community levels, we calculate the proportions of other ROIs in the same community that belong to each of the 18 reference systems. We then average these proportions across the three levels to obtain an 18-dimensional affiliation vector for each ROI. The resulting profile describes each ROI's community composition using a consistent reference frame across participants.

We subtract the mean profile of the training participants from each participant's profile and use the resulting flattened features as input to a regularized logistic regression model (Supplementary Section~\ref{sup:affil}). Using the affiliation profile alone, the model achieves an AUROC of 0.707 on ABIDE and 0.725 on ADNI on held-out test participants. Compared with a model trained using covariates alone (0.602 and 0.693), adding the affiliation profile to the covariates improves AUROC by 0.079 and 0.041, respectively, to 0.681 and 0.734. The covariates include age, sex, head motion, and the number and size concentration of the communities at each level, as well as scanner vendor, repetition time, and scan length for ADNI. In this combined model, when each participant's profile is replaced by that of another randomly selected participant from the same site, the AUROC drops back to the covariate-only level (0.597 and 0.689). These results suggest that individualized community assignments provide diagnostic information beyond the measured covariates, and that this additional information depends on each participant's own community profile.

\subsection{Interpretability}
\label{sec:disease-communities}
In this section, we explore how the individualized communities differ between patients and healthy controls, and whether IC-GNN reflects these differences. We first identify the differences in community organization associated with diagnosis using the communities generated by AP. For each participant, we first measure how frequently ROIs within or between functional systems share a community. Specifically, we compute the proportion of ROI pairs assigned to the same first-level community and correct it for chance co-assignment (Supplementary Section~\ref{sup:interp-org}).

\begin{figure*}[t]
\centering
\includegraphics[width=0.66\textwidth]{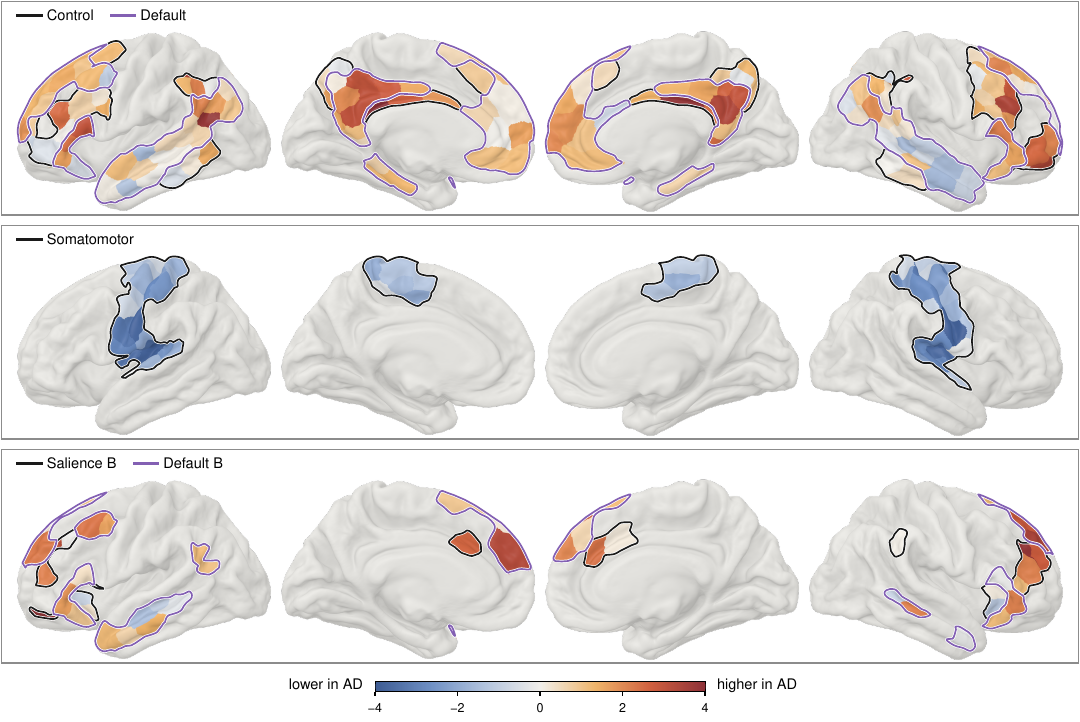}
\caption{Differences in community organization between AD and CN participants. For each ROI, we compare how often it shares a community with ROIs of a partner network between AD and CN participants: control with default (top), somatomotor with somatomotor (middle), and salience B with default B (bottom). Colors show the $t$ statistic of the AD vs.\ CN difference from a regression analysis adjusted for covariates. Red indicates that the ROI is grouped with its partner network more often in AD, and blue means less often. The outlines mark these networks.}
\label{fig:ad-cn}
\end{figure*}

In both ABIDE and ADNI, the control and default networks are grouped more often in ASD and AD patients ($p=0.008$ and $p=0.023$). In AD, this increase is most pronounced in the posterior cingulate and precuneus (Fig.~\ref{fig:ad-cn}), which are major cortical hubs~\citep{Buckner2009} and among the earliest sites of amyloid accumulation~\citep{Palmqvist2017}. This may also suggest that the boundary between the control and default networks becomes less distinct in these regions. Salience B and default B ROIs, mainly in the prefrontal cortex, are also grouped with each other more often in AD ($p=0.020$). These results are consistent with the reduced segregation between the control and default networks in AD~\citep{Contreras2019} and the hyperconnectivity between the default network and the rest of the brain in ASD~\citep{Ilioska2023}. On the other hand, somatomotor ROIs are grouped together less often in ASD ($p=0.005$). In ADNI, this difference keeps its direction ($p=0.055$) and is strongest in the ventral somatomotor cortex. This aligns with the involvement of sensory-motor systems in network changes associated with dementia severity~\citep{Zhang2023}.

\begin{figure*}[t]
\centering
\includegraphics[width=0.92\textwidth]{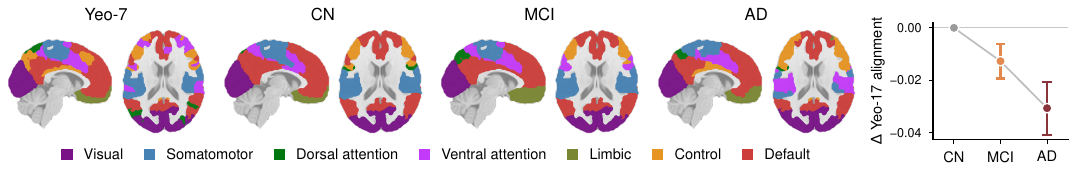}
\caption{Community organization across CN, MCI, and AD. The maps show the Yeo-7 networks and the group-consensus communities in ADNI, each colored by its majority Yeo-7 network. The right panel shows the alignment with Yeo-17 relative to CN, defined as the average fraction of an ROI's community members that belong to the same Yeo-17 network as the ROI, with 95\% confidence intervals.}
\label{fig:consensus}
\end{figure*}

We next compare each participant's overall community partition with the Yeo-17 atlas using NMI. We find that the communities of patients show less alignment with the atlas in both datasets ($p=0.012$ and $p=0.0068$). This indicates a reduced correspondence between individual community structure and the canonical networks, as reported in ASD~\citep{Keown2017}. In the held-out ADNI sites, all of these community differences are graded from CN to MCI to AD (one-sided trend $p\le0.005$). Across all ADNI participants, the alignment with the Yeo-17 atlas is lower in MCI than in CN and lowest in AD (Fig.~\ref{fig:consensus}), and the decrease in AD is seen across functional systems, suggesting the progressive disruption of canonical networks in AD.

\begin{figure*}[t]
\centering
\includegraphics[width=0.92\textwidth]{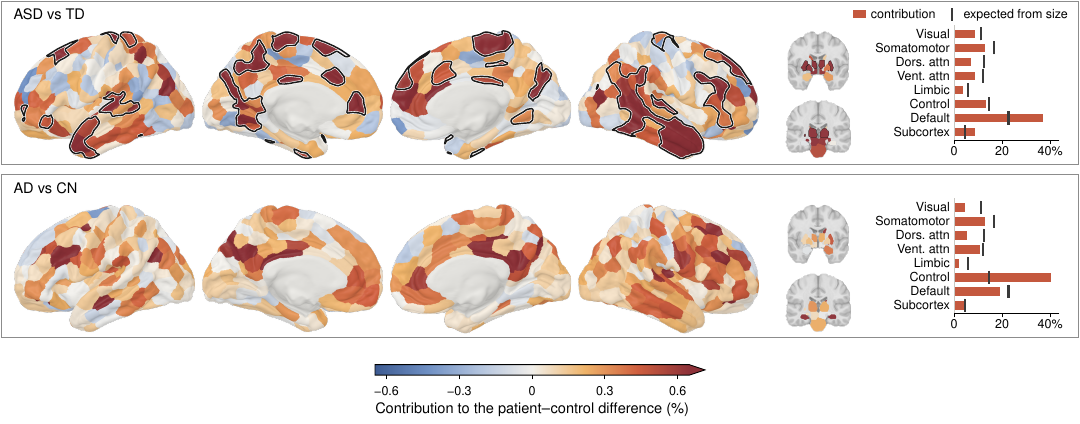}
\caption{Regional evidence differences between patients and controls. Brain maps show each ROI's contribution to the difference in model logits (without the intercept) between patients and controls, expressed as a percentage of the total difference. Positive values indicate higher regional evidence in patients, while negative values indicate higher evidence in controls. Results are shown for ASD versus typically developing controls (ABIDE) and AD versus cognitively normal controls (ADNI). Two coronal slices show subcortical regions. Black outlines mark significant regional differences in ASD. The bars sum the contributions of the ROIs in each functional network, and the ticks indicate each network's proportion of the 419 ROIs. Results are based on held-out test data and averaged across training seeds.}
\label{fig:evidence}
\end{figure*}

We then test whether the regional evidence that IC-GNN uses for prediction varies with the differences in community organization across held-out participants. Our design in Section~\ref{sec:readout} provides a natural way to analyze this by examining the regional evidence $c_i$ of each ROI. Since IC-GNN decomposes its prediction logit into contributions from 419 ROIs, we aggregate these contributions by functional system (Supplementary Section~\ref{sup:interp-model}). On ADNI, the control network accounts for 40\% of the difference between AD and CN in the summed evidence, mostly from regions in the posterior cingulate and precuneus, where the grouping of control and default ROIs increases most. On ABIDE, the default network accounts for 37\% of the difference between ASD and controls, mostly from the lateral temporal cortex and the dorsal and medial prefrontal cortex (Fig.~\ref{fig:evidence}). These regions, as well as the thalamus and basal ganglia, which also contribute more than their share of ROIs, have been implicated in the ASD literature~\citep{Zilbovicius2006,Padmanabhan2017,Cerliani2015}. In addition, we perform a regression analysis between the co-assignment of the control and default networks and the summed evidence of their ROIs across held-out ADNI participants. We find that the more the default ROIs of a participant are grouped with the control ROIs, the more evidence for AD the model reads from the two networks (standardized $\beta=0.18$, permutation $p<0.001$), and this association also holds among participants with the same diagnosis ($\beta=0.13$, $p<0.001$). This indicates that the model tracks continuous individual variations in the blurring of the control and default networks, even within the same diagnostic group. On ABIDE, we also observe a similar trend ($\beta=0.05$, $p=0.12$). These results suggest that the regional evidence learned by IC-GNN reflects individual differences in community organization.

\section{Discussion}
\label{sec:discussion}

The brain exhibits a modular functional organization, and previous studies have shown that differences in this organization are associated with cognition and neurological or psychiatric disorders~\citep{Kong2019,Kong2021,ad-meta-review,Zhang2023,Lynch2024}. Recent studies have increasingly recognized individualized parcellation as a complement to group-level atlases because it captures individual differences in functional organization that are not fully represented by group-level brain parcellations~\citep{Hermosillo2024}. However, fine-grained individualized parcellation at the voxel level is sensitive to fMRI data quality and complicates cross-subject alignment of functional communities. We adopt another paradigm that builds on individualized parcellation at a different granularity, the atlas-anchored region level. By dynamically aggregating fine-grained atlas regions into functional communities of varying sizes and scales, we establish a data-efficient scheme that preserves cross-subject alignment while capturing individualized community organization. Grounded in this concept, we developed IC-GNN, a self-explainable framework that utilizes individualized multiscale communities to guide graph representation learning and adapt ROI-specific prediction weights. Within this framework, we demonstrate the value of subject-specific community organization both as a robust disease-associated biomarker and as an interpretable tool for neuroscientific discovery and precision medicine.

IC-GNN was evaluated across three datasets covering ASD, AD, and depression, and achieved the highest mean AUROC and accuracy among the compared methods. Beyond its superior benchmark performance, its architectural design also aligns closely with intrinsic brain network organization. Specifically, our ablation analyses reveal three key insights into modeling disordered brain networks. First, removing the community hierarchy substantially degraded performance, highlighting that capturing multiscale interactions is critical for modeling complex brain disorders. More importantly, replacing subject-specific communities with consensus or random partitions also reduced performance on ABIDE, which suggests the importance of integrating individualized topological variations that a shared partition cannot capture. Finally, the gains from learned edge refinement demonstrate the strength of incorporating subject-specific information to guide fMRI connectivity denoising and ensure robust graph representation learning.

We then examined whether individualized communities capture reproducible individual differences while remaining consistent with known functional organization. We found that the communities contain individual-specific patterns that are reproducible both within a single scan and across scanning sessions. Although community assignments differed among participants, they showed considerable alignment with the Yeo-17 atlas. This alignment became stronger when the communities were aggregated across participants into a group-level consensus. These findings suggest that individualized communities reflect shared functional organization while preserving reproducible individual differences that a group-level parcellation cannot represent.

We then investigated whether these individual differences are also associated with brain disorders. Community affiliation profiles alone showed predictive ability for ASD and AD, and provided additional information beyond the measured demographic and acquisition covariates. This information was lost when a participant's profile was replaced by that of another participant from the same site. We also identified disease-associated changes in community organization in both ABIDE and ADNI, including increased co-assignment between control and default network regions in both cohorts and reduced within-system grouping of somatomotor regions in ABIDE. In held-out ADNI sites, these differences were graded from CN to MCI to AD. Therefore, individual differences in community organization not only encode personal characteristics but also contain information relevant to brain disorders.

Finally, we found that these disease-associated changes are reflected in the regional evidence produced by the community-guided readout. In ADNI, greater co-assignment between the control and default networks was associated with stronger evidence for AD from these networks, even among participants with the same diagnosis. This relationship suggests that the regional evidence learned by IC-GNN reflects individual differences in functional community organization.

In summary, individualized communities capture reproducible, disease-related individual differences, and IC-GNN incorporates them into an accurate, self-explainable model that provides traceable regional evidence. These findings support the value of explicitly modeling individual functional organization for brain network analysis.

\section*{Acknowledgments and funding}
The authors thank the participants and investigators of the Autism Brain Imaging Data Exchange (ABIDE), the Alzheimer's Disease Neuroimaging Initiative (ADNI), and the UK Biobank for making the data used in this study available. The authors also gratefully acknowledge the Yale Center for Research Computing for use of the research computing infrastructure.

For ABIDE, primary support for the work by Adriana Di Martino was provided by the NIMH (K23MH087770) and the Leon Levy Foundation. Primary support for the work by Michael P. Milham and the INDI team was provided by gifts from Joseph P. Healy and the Stavros Niarchos Foundation to the Child Mind Institute, as well as by an NIMH award to MPM (NIMH R03MH096321).

Data collection and sharing for the Alzheimer's Disease Neuroimaging Initiative (ADNI) is funded by the National Institute on Aging (National Institutes of Health Grant U19AG024904). The grantee organization is the Northern California Institute for Research and Education. In the past, ADNI has also received funding from the National Institute of Biomedical Imaging and Bioengineering, the Canadian Institutes of Health Research, and private sector contributions through the Foundation for the National Institutes of Health (FNIH) including generous contributions from the following: AbbVie, Alzheimer's Association; Alzheimer's Drug Discovery Foundation; Araclon Biotech; BioClinica, Inc.; Biogen; Bristol-Myers Squibb Company; CereSpir, Inc.; Cogstate; Eisai Inc.; Elan Pharmaceuticals, Inc.; Eli Lilly and Company; EuroImmun; F. Hoffmann-La Roche Ltd and its affiliated company Genentech, Inc.; Fujirebio; GE Healthcare; IXICO Ltd.; Janssen Alzheimer Immunotherapy Research \& Development, LLC.; Johnson \& Johnson Pharmaceutical Research \& Development LLC.; Lumosity; Lundbeck; Merck \& Co., Inc.; Meso Scale Diagnostics, LLC.; NeuroRx Research; Neurotrack Technologies; Novartis Pharmaceuticals Corporation; Pfizer Inc.; Piramal Imaging; Servier; Takeda Pharmaceutical Company; and Transition Therapeutics.

This research has been conducted using the UK Biobank Resource under Application Number 41170. The authors thank all UK Biobank participants and the staff involved in study design, recruitment, data collection, and data dissemination.

This research did not receive any specific grant from funding agencies in the public, commercial, or not-for-profit sectors.

\section*{Ethics and data access}
This study is a secondary analysis of de-identified data from ABIDE, ADNI, and UK Biobank. ABIDE data contributions were based on studies approved by local institutional review boards, and all data were fully anonymized before release~\citep{ABIDE2014}. ADNI was approved by the institutional review boards of all participating institutions, and written informed consent was obtained from all participants or their authorized representatives. UK Biobank has approval from the North West Multi-centre Research Ethics Committee as a Research Tissue Bank (REC reference 21/NW/0157), under which approved researchers can use the resource without separate ethical approval; all UK Biobank participants provided written informed consent. All data were accessed under the data use agreements of the respective providers.

\section*{Declaration of competing interest}
The authors declare that they have no known competing financial interests or personal relationships that could have appeared to influence the work reported in this paper.

\section*{Data and code availability}
ABIDE, ADNI, and UK Biobank data are available under the access conditions of their respective providers. Participant-level data and derived subject files remain subject to those conditions. The code will be made publicly available upon publication.

\bibliographystyle{elsarticle-harv}
\bibliography{references}

\clearpage
\onecolumn
\pdfbookmark[0]{Supplementary Material}{supplementary-material}
\setcounter{section}{0}\setcounter{subsection}{0}\setcounter{table}{0}\setcounter{figure}{0}\setcounter{equation}{0}
\renewcommand{\thesection}{S\arabic{section}}
\renewcommand{\thetable}{S\arabic{table}}
\renewcommand{\thefigure}{S\arabic{figure}}
\renewcommand{\theequation}{S\arabic{equation}}
\def\theHsection{S\arabic{section}}
\def\theHtable{S\arabic{table}}
\def\theHfigure{S\arabic{figure}}
\def\theHequation{S\arabic{equation}}
\begin{center}
{\LARGE Supplementary Material}
\end{center}
\vspace{1em}
\section{Complete benchmark results}\label{sup:bench}

\subsection{Metrics}\label{sup:bench-protocol}

For threshold-dependent metrics, the threshold of each model is selected on the validation set of each fold and applied to the corresponding test set. On ABIDE and ADNI, we use the threshold that maximizes Youden's index (sensitivity $+$ specificity $-$~1). On UK Biobank, all models are compared at a common operating point, the threshold that targets a 37.5\% positive-call rate on the validation set.

The disease group (ASD, AD or depression) is the positive class, and AUROC is computed from the continuous model outputs. With true positives ($TP$), false positives ($FP$), true negatives ($TN$) and false negatives ($FN$) counted on the test participants of a fold, we compute six threshold-dependent metrics: accuracy, recall (sensitivity), specificity, balanced accuracy (BAcc), precision and $F_1$ score.
\begin{align*}
\mathrm{Accuracy} &= \frac{TP+TN}{TP+FP+TN+FN}, & \mathrm{Recall} &= \frac{TP}{TP+FN}, & \mathrm{Specificity} &= \frac{TN}{TN+FP},\\
\mathrm{BAcc} &= \frac{\mathrm{Recall}+\mathrm{Specificity}}{2}, & \mathrm{Precision} &= \frac{TP}{TP+FP}, & F_1 &= \frac{2\,TP}{2\,TP+FP+FN}.
\end{align*}

\subsection{Results}\label{sup:bench-results}

Tables~\ref{tab:sup-bench-abide}--\ref{tab:sup-bench-ukb} report all metrics for every model. The AUROC and accuracy columns are those of Table~\ref{tab:accuracy_three_datasets}. IC-GNN achieves the highest accuracy and specificity on all three datasets, the highest balanced accuracy and $F_1$ score on ABIDE and ADNI, and the highest precision on ADNI and UK Biobank.

\begin{table}[htbp]
\centering
\caption{Benchmark results on ABIDE. Values are percentages (mean $\pm$ SD over the five test folds).}
\label{tab:sup-bench-abide}
\resizebox{\textwidth}{!}{%
\begin{tabular}{@{}lccccccc@{}}
\toprule
Model & AUROC & Accuracy & BAcc & Recall & Specificity & Precision & $F_1$ \\
\midrule
\multicolumn{8}{@{}l}{\textit{Conventional ML}} \\
SVM & $71.42 \pm 5.34$ & $63.49 \pm 2.42$ & $63.06 \pm 3.65$ & $62.30 \pm 18.28$ & $63.82 \pm 12.64$ & $60.50 \pm 5.24$ & $60.00 \pm 9.93$ \\
Random Forest & $65.85 \pm 4.96$ & $61.73 \pm 3.14$ & $60.56 \pm 3.41$ & $55.88 \pm 19.61$ & $65.24 \pm 13.23$ & $58.80 \pm 3.38$ & $55.76 \pm 11.65$ \\
\midrule
\multicolumn{8}{@{}l}{\textit{General-purpose GNNs}} \\
GCN & $61.46 \pm 5.61$ & $58.60 \pm 3.76$ & $57.64 \pm 5.18$ & $52.68 \pm 22.55$ & $62.60 \pm 13.86$ & $55.10 \pm 8.77$ & $51.15 \pm 15.30$ \\
GAT & $62.97 \pm 5.98$ & $58.52 \pm 3.43$ & $57.75 \pm 4.23$ & $59.99 \pm 17.04$ & $55.50 \pm 14.02$ & $55.24 \pm 8.70$ & $55.63 \pm 10.77$ \\
GIN & $56.54 \pm 4.34$ & $52.34 \pm 5.98$ & $52.27 \pm 4.92$ & $49.04 \pm 18.85$ & $55.49 \pm 22.49$ & $51.22 \pm 8.69$ & $46.08 \pm 10.73$ \\
GraphSAGE & $62.53 \pm 5.36$ & $59.51 \pm 5.01$ & $59.39 \pm 4.72$ & $63.33 \pm 14.86$ & $55.45 \pm 15.94$ & $56.51 \pm 5.38$ & $58.18 \pm 7.72$ \\
\midrule
\multicolumn{8}{@{}l}{\textit{Brain-network models}} \\
BrainNetTF & $\underline{73.84} \pm 6.19$ & $66.41 \pm 4.70$ & $64.77 \pm 5.97$ & $62.04 \pm 23.06$ & $67.50 \pm 21.20$ & $68.16 \pm 11.90$ & $60.41 \pm 14.71$ \\
BrainNetCNN & $70.91 \pm 5.15$ & $65.74 \pm 3.83$ & $64.19 \pm 3.76$ & $63.38 \pm 20.15$ & $65.01 \pm 19.06$ & $64.69 \pm 6.55$ & $61.25 \pm 11.01$ \\
BrainGNN & $59.18 \pm 5.24$ & $55.30 \pm 3.49$ & $55.80 \pm 2.40$ & $\underline{64.32} \pm 18.15$ & $47.27 \pm 17.20$ & $52.49 \pm 6.35$ & $55.94 \pm 8.69$ \\
ASD-DiagNet & $71.74 \pm 7.74$ & $\underline{66.98} \pm 4.77$ & $\underline{65.35} \pm 5.88$ & $\mathbf{65.90} \pm 21.18$ & $64.81 \pm 21.83$ & $65.29 \pm 8.77$ & $\underline{63.24} \pm 12.07$ \\
BQN & $68.24 \pm 5.48$ & $63.55 \pm 4.63$ & $62.54 \pm 4.64$ & $62.75 \pm 20.21$ & $62.33 \pm 21.63$ & $62.70 \pm 6.53$ & $60.05 \pm 9.90$ \\
Com-BrainTF & $70.38 \pm 8.04$ & $66.03 \pm 4.81$ & $64.49 \pm 6.38$ & $59.01 \pm 19.28$ & $69.98 \pm 13.15$ & $63.69 \pm 8.34$ & $59.56 \pm 15.11$ \\
BrainHGT & $69.48 \pm 8.00$ & $61.98 \pm 3.19$ & $59.34 \pm 5.02$ & $48.75 \pm 27.66$ & $69.94 \pm 21.83$ & $60.00 \pm 11.54$ & $49.75 \pm 19.53$ \\
ALTER & $70.73 \pm 5.19$ & $66.20 \pm 2.14$ & $63.89 \pm 1.24$ & $54.06 \pm 20.86$ & $\underline{73.71} \pm 20.83$ & $\mathbf{70.22} \pm 7.58$ & $57.76 \pm 10.06$ \\
CAGT & $65.91 \pm 5.94$ & $62.61 \pm 2.30$ & $60.27 \pm 2.08$ & $50.38 \pm 20.89$ & $70.16 \pm 18.44$ & $62.60 \pm 2.53$ & $53.33 \pm 12.05$ \\
\midrule
\multicolumn{8}{@{}l}{\textit{Ours}} \\
\textbf{IC-GNN} & $\mathbf{75.41} \pm 5.73$ & $\mathbf{69.40} \pm 4.23$ & $\mathbf{67.82} \pm 4.89$ & $60.22 \pm 15.58$ & $\mathbf{75.43} \pm 12.77$ & $\underline{70.04} \pm 4.87$ & $\mathbf{63.55} \pm 9.72$ \\
\bottomrule
\end{tabular}%
}
\end{table}
\begin{table}[htbp]
\centering
\caption{Benchmark results on ADNI. Values are percentages (mean $\pm$ SD over the five test folds).}
\label{tab:sup-bench-adni}
\resizebox{\textwidth}{!}{%
\begin{tabular}{@{}lccccccc@{}}
\toprule
Model & AUROC & Accuracy & BAcc & Recall & Specificity & Precision & $F_1$ \\
\midrule
\multicolumn{8}{@{}l}{\textit{Conventional ML}} \\
SVM & $82.54 \pm 2.21$ & $71.22 \pm 4.38$ & $75.50 \pm 2.73$ & $\mathbf{82.14} \pm 6.47$ & $68.87 \pm 6.16$ & $36.72 \pm 3.48$ & $50.51 \pm 2.59$ \\
Random Forest & $70.25 \pm 4.40$ & $65.64 \pm 6.96$ & $65.40 \pm 2.53$ & $65.02 \pm 16.10$ & $65.79 \pm 12.03$ & $29.82 \pm 2.73$ & $39.79 \pm 1.89$ \\
\midrule
\multicolumn{8}{@{}l}{\textit{General-purpose GNNs}} \\
GCN & $74.89 \pm 2.90$ & $69.52 \pm 8.51$ & $66.82 \pm 2.66$ & $62.56 \pm 12.58$ & $71.08 \pm 12.72$ & $34.01 \pm 7.06$ & $42.63 \pm 4.48$ \\
GAT & $77.59 \pm 3.33$ & $73.02 \pm 7.76$ & $68.87 \pm 5.50$ & $62.39 \pm 18.89$ & $75.35 \pm 12.66$ & $37.12 \pm 4.81$ & $44.53 \pm 5.80$ \\
GIN & $71.98 \pm 2.99$ & $66.36 \pm 5.17$ & $65.04 \pm 2.81$ & $62.87 \pm 9.81$ & $67.22 \pm 7.73$ & $29.61 \pm 3.67$ & $39.84 \pm 3.84$ \\
GraphSAGE & $76.54 \pm 1.41$ & $68.93 \pm 7.32$ & $67.97 \pm 4.41$ & $66.66 \pm 19.55$ & $69.27 \pm 12.79$ & $32.89 \pm 2.46$ & $42.64 \pm 4.03$ \\
\midrule
\multicolumn{8}{@{}l}{\textit{Brain-network models}} \\
BrainNetTF & $83.98 \pm 1.81$ & $\underline{78.56} \pm 6.06$ & $75.51 \pm 3.29$ & $70.87 \pm 8.73$ & $\underline{80.14} \pm 8.54$ & $\underline{45.54} \pm 7.88$ & $54.56 \pm 5.49$ \\
BrainNetCNN & $81.20 \pm 4.59$ & $71.69 \pm 7.04$ & $72.08 \pm 3.60$ & $72.76 \pm 6.44$ & $71.41 \pm 9.38$ & $36.56 \pm 5.36$ & $48.28 \pm 4.75$ \\
BrainGNN & $75.13 \pm 1.51$ & $65.24 \pm 7.02$ & $68.02 \pm 1.53$ & $72.43 \pm 8.43$ & $63.62 \pm 10.39$ & $30.74 \pm 3.00$ & $42.77 \pm 1.77$ \\
ASD-DiagNet & $\underline{84.78} \pm 3.59$ & $77.29 \pm 5.30$ & $\underline{76.68} \pm 1.90$ & $75.77 \pm 5.87$ & $77.59 \pm 7.48$ & $44.19 \pm 9.12$ & $\underline{54.86} \pm 4.72$ \\
BQN & $79.50 \pm 3.38$ & $71.54 \pm 5.81$ & $72.38 \pm 2.89$ & $73.75 \pm 5.72$ & $71.02 \pm 7.98$ & $36.32 \pm 4.76$ & $48.30 \pm 3.65$ \\
Com-BrainTF & $82.29 \pm 5.85$ & $74.88 \pm 6.53$ & $74.47 \pm 5.95$ & $73.91 \pm 6.38$ & $75.03 \pm 7.08$ & $39.97 \pm 6.76$ & $51.71 \pm 7.04$ \\
BrainHGT & $84.51 \pm 2.84$ & $75.14 \pm 6.41$ & $75.64 \pm 3.07$ & $\underline{76.37} \pm 10.71$ & $74.90 \pm 9.63$ & $41.33 \pm 6.69$ & $52.68 \pm 3.85$ \\
ALTER & $84.10 \pm 2.72$ & $74.28 \pm 6.91$ & $74.67 \pm 3.38$ & $75.50 \pm 11.87$ & $73.84 \pm 10.39$ & $40.54 \pm 7.05$ & $51.65 \pm 4.11$ \\
CAGT & $80.90 \pm 2.89$ & $76.28 \pm 5.07$ & $72.28 \pm 2.45$ & $66.18 \pm 9.40$ & $78.39 \pm 7.87$ & $40.92 \pm 4.27$ & $49.97 \pm 2.40$ \\
\midrule
\multicolumn{8}{@{}l}{\textit{Ours}} \\
\textbf{IC-GNN} & $\mathbf{86.46} \pm 3.03$ & $\mathbf{80.52} \pm 5.16$ & $\mathbf{78.00} \pm 2.91$ & $74.11 \pm 7.28$ & $\mathbf{81.89} \pm 7.10$ & $\mathbf{48.84} \pm 8.43$ & $\mathbf{58.04} \pm 5.19$ \\
\bottomrule
\end{tabular}%
}
\end{table}
\begin{table}[htbp]
\centering
\caption{Benchmark results on UK Biobank. Values are percentages (mean $\pm$ SD over the five folds).}
\label{tab:sup-bench-ukb}
\resizebox{\textwidth}{!}{%
\begin{tabular}{@{}lccccccc@{}}
\toprule
Model & AUROC & Accuracy & BAcc & Recall & Specificity & Precision & $F_1$ \\
\midrule
\multicolumn{8}{@{}l}{\textit{Conventional ML}} \\
SVM & $59.06 \pm 0.28$ & $61.70 \pm 0.53$ & $55.98 \pm 0.23$ & $46.37 \pm 0.57$ & $\underline{65.58} \pm 0.77$ & $25.43 \pm 0.27$ & $32.84 \pm 0.22$ \\
Random Forest & $51.69 \pm 0.28$ & $57.01 \pm 1.11$ & $50.85 \pm 0.31$ & $40.51 \pm 1.23$ & $61.19 \pm 1.69$ & $20.90 \pm 0.27$ & $27.56 \pm 0.21$ \\
\midrule
\multicolumn{8}{@{}l}{\textit{General-purpose GNNs}} \\
GCN & $57.04 \pm 0.46$ & $60.35 \pm 0.40$ & $54.26 \pm 0.32$ & $44.04 \pm 1.06$ & $64.48 \pm 0.69$ & $23.88 \pm 0.25$ & $30.96 \pm 0.43$ \\
GAT & $57.12 \pm 0.11$ & $60.83 \pm 0.20$ & $54.97 \pm 0.51$ & $45.13 \pm 1.35$ & $64.81 \pm 0.43$ & $24.49 \pm 0.39$ & $31.75 \pm 0.65$ \\
GIN & $55.52 \pm 0.62$ & $59.99 \pm 0.54$ & $53.52 \pm 0.30$ & $42.68 \pm 0.69$ & $64.37 \pm 0.78$ & $23.26 \pm 0.28$ & $30.11 \pm 0.30$ \\
GraphSAGE & $57.60 \pm 0.31$ & $60.41 \pm 0.54$ & $54.81 \pm 0.47$ & $45.43 \pm 0.80$ & $64.20 \pm 0.69$ & $24.31 \pm 0.43$ & $31.67 \pm 0.50$ \\
\midrule
\multicolumn{8}{@{}l}{\textit{Brain-network models}} \\
BrainNetTF & $54.13 \pm 1.75$ & $59.36 \pm 0.82$ & $52.39 \pm 1.03$ & $40.69 \pm 1.86$ & $64.08 \pm 0.91$ & $22.28 \pm 0.88$ & $28.79 \pm 1.17$ \\
BrainNetCNN & $58.04 \pm 0.64$ & $60.86 \pm 0.77$ & $55.41 \pm 0.79$ & $46.28 \pm 1.03$ & $64.55 \pm 0.81$ & $24.83 \pm 0.72$ & $32.32 \pm 0.83$ \\
BrainGNN & $56.98 \pm 0.33$ & $59.96 \pm 0.69$ & $54.20 \pm 0.49$ & $44.55 \pm 0.99$ & $63.86 \pm 0.97$ & $23.78 \pm 0.45$ & $31.00 \pm 0.52$ \\
ASD-DiagNet & $58.07 \pm 0.47$ & $60.80 \pm 0.58$ & $55.17 \pm 0.58$ & $45.73 \pm 1.31$ & $64.61 \pm 0.82$ & $24.64 \pm 0.50$ & $32.02 \pm 0.67$ \\
BQN & $57.68 \pm 0.77$ & $61.58 \pm 0.75$ & $55.73 \pm 0.70$ & $45.91 \pm 1.11$ & $65.54 \pm 0.91$ & $25.22 \pm 0.66$ & $32.55 \pm 0.76$ \\
Com-BrainTF & $\underline{60.35} \pm 0.56$ & $\underline{62.00} \pm 0.50$ & $\mathbf{57.28} \pm 0.55$ & $\mathbf{49.36} \pm 1.21$ & $65.19 \pm 0.70$ & $\underline{26.41} \pm 0.47$ & $\mathbf{34.40} \pm 0.62$ \\
BrainHGT & $59.34 \pm 0.78$ & $61.39 \pm 0.42$ & $56.18 \pm 0.40$ & $47.44 \pm 0.53$ & $64.92 \pm 0.48$ & $25.49 \pm 0.37$ & $33.16 \pm 0.42$ \\
ALTER & $60.10 \pm 0.43$ & $61.94 \pm 0.69$ & $56.71 \pm 0.62$ & $47.96 \pm 0.87$ & $65.47 \pm 0.84$ & $26.01 \pm 0.58$ & $33.72 \pm 0.66$ \\
CAGT & $59.16 \pm 0.66$ & $61.89 \pm 0.94$ & $56.84 \pm 0.98$ & $48.36 \pm 1.26$ & $65.31 \pm 0.98$ & $26.09 \pm 0.92$ & $33.89 \pm 1.05$ \\
\midrule
\multicolumn{8}{@{}l}{\textit{Ours}} \\
\textbf{IC-GNN} & $\mathbf{60.62} \pm 0.45$ & $\mathbf{62.51} \pm 0.67$ & $\underline{57.25} \pm 0.53$ & $\underline{48.42} \pm 0.73$ & $\mathbf{66.08} \pm 0.84$ & $\mathbf{26.54} \pm 0.54$ & $\underline{34.28} \pm 0.56$ \\
\bottomrule
\end{tabular}%
}
\end{table}

\section{Ablation experiments}\label{sup:abl}

In this section, we elaborate on the details of each ablation setting in the main paper. Each ablated model in Table~\ref{tab:ablation_abide} modifies IC-GNN as described below and is trained from scratch on ABIDE with the same folds, training seeds and training settings as the full model.
\begin{itemize}
\item \emph{Hierarchy removal}: the model without the hierarchical community structure. The encoder keeps only the first graph convolution and its edge refinement, which no longer uses community membership, and the readout keeps only the ROI level.
\item \emph{Random community control}: each participant's communities are replaced by random communities obtained by randomly permuting their community assignments across the nodes of each level. The number and sizes of communities remain the same.
\item \emph{Consensus community control}: all participants share one community organization at each level. The training participants' mean co-assignment matrix is clustered by average linkage and cut at the median number of communities at each level. Each participant keeps their own graph.
\item \emph{Open edge control}: the probability matrix obtained from PERM at every level is replaced by a fixed probability of 0.999, so that all edges are kept.
\item \emph{Uniform edge control}: the probability matrix obtained from PERM at every level is replaced by uniform attenuation at the keep rate learned by the full model at that level, so that the overall attenuation is preserved but individual edges are no longer selected.
\item \emph{Joint control}: hierarchy removal combined with open edges.
\end{itemize}

\section{Details of the community analyses}\label{sup:methods}

The analyses in this section use each participant's individualized community hierarchy (Section~\ref{sec:ap}). Our reference systems consist of the 17 Yeo networks~\citep{Yeo2011} for the 400 cortical ROIs and one additional system for the 19 noncortical ROIs; where stated, the 17 cortical networks are further grouped into the 7 canonical Yeo networks. Across the three community levels, the median numbers of communities are 79, 29, and 11 for ABIDE; 83, 31, and 12 for ADNI; and 89, 34, and 13 for UK Biobank.

\subsection{Individualized communities}\label{sup:comm}

\paragraph{Identification} On ABIDE, we split each participant's regional time series into two contiguous halves of equal length and compute a first-level partition using AP for each half. On ADNI, for the 980 participants with a follow-up resting-state session, we compute partitions from the primary analyzed scan and the nearest other available resting-state session (median interval 14 months). In both datasets, this results in two partitions per participant (the two halves on ABIDE, or the two sessions on ADNI). We then measure the similarity between any two first-level partitions $P$ and $P'$ using the Jaccard index of their co-assigned ROI pairs:
\begin{equation}
J(P,P')=\frac{\lvert\mathcal{C}(P)\cap\mathcal{C}(P')\rvert}{\lvert\mathcal{C}(P)\cup\mathcal{C}(P')\rvert},
\end{equation}
where $\mathcal{C}(P)$ is the set of ROI pairs assigned to the same community in $P$. To test whether these community partitions can identify individuals within each site, we search the second partitions of all participants from the same site for the one most similar to each participant's first partition. A participant is correctly identified if their own second partition is the single most similar match. The chance level is the mean of $1/n_s$ across participants, where $n_s$ is the number of participants at a given site (2.3\% for ABIDE and 6.3\% for ADNI). In the control setting, we replace each participant's first partition with a random partition with the same community sizes and repeat the search, which results in a similar performance to the chance level (3.8\% for ABIDE and 6.3\% for ADNI).

\paragraph{Alignment with the atlas} After obtaining the first-level partition $U$ of each participant using AP, we compute its normalized mutual information (NMI) with the reference partition $V$ (17 Yeo networks and one noncortical system):
\begin{equation}
\mathrm{NMI}(U,V)=\frac{I(U;V)}{\tfrac{1}{2}\left[H(U)+H(V)\right]},
\end{equation}
where $I$ is the mutual information and $H$ is the entropy of the label proportions. The mean NMI is 0.571 for ABIDE, 0.535 for ADNI, and 0.557 for UK Biobank, compared with about 0.35 for random communities of the same sizes. We also average the co-assignment matrices across all participants and apply agglomerative hierarchical clustering to obtain a consensus partition matched to the cohort's median number of first-level communities. The NMI increases to 0.700, 0.694, and 0.726, respectively.

\subsection{Diagnostic information in individualized communities}\label{sup:affil}

To test whether individualized communities carry diagnostic information, we constructed an affiliation profile for each participant. Let $a^{(l)}(i)$ denote the community assignment of ROI $i$ at hierarchy level $l \in \{1,2,3\}$, $n^{(l)}_{i}$ the size of this community, and $g(i)$ the canonical reference system of ROI $i$. The affiliation of ROI $i$ to system $k$ is defined as:
\begin{equation}
\phi_{i,k} = \frac{1}{3}\sum_{l=1}^{3}
\frac{\bigl|\{\,j \neq i:\ a^{(l)}(j) = a^{(l)}(i),\ g(j) = k\,\}\bigr|}{\max\bigl(n^{(l)}_{i}-1,\,1\bigr)},
\qquad k=1,\dots,18,
\end{equation}
which quantifies the average fraction of community co-members of ROI $i$ that belong to reference system $k$ across the three levels. Each participant is thus characterized by a $419 \times 18$ matrix (7,542 features), centered by subtracting the feature-wise training mean.

To eliminate site-related bias, we fitted L2-regularized logistic regression models using five-fold site-grouped cross-validation under three feature configurations: covariates only, profile features only, and both combined.

The baseline model adjusted for potential demographic, motion, topological, and acquisition confounds:
\begin{itemize}
    \item \textbf{Demographics:} Age modeled using a natural cubic spline with four degrees of freedom to capture non-linear aging trajectories, and sex.
    \item \textbf{Head motion:} The natural logarithm of mean framewise displacement ($\ln \overline{\text{FD}}$).
    \item \textbf{Global partition properties:} At each level $l$, the total number of detected cortical communities and their size concentration, quantified by the Herfindahl-like index $\sum_c (n_c / 400)^2$, where $n_c$ is the number of cortical ROIs in community $c$.
    \item \textbf{Imaging acquisition (ADNI only):} Scanner vendor, a binary indicator for sub-second repetition time ($\text{TR} < 1\text{ s}$), and the natural logarithm of the number of recorded volumes.
\end{itemize}

Classification performance was evaluated by test AUROC averaged across folds.

To confirm that the diagnostic performance reflected individual community organization instead of site-level artifacts, we conducted a null control experiment by permuting profiles within sites. Specifically, within each fold, data split, and acquisition site, the affiliation profiles were randomly shuffled across participants independently of diagnostic labels. Under this scheme, each participant retained their true diagnosis and covariates but was assigned the profile of another individual from the same scanning site. The combined model (covariates plus profile) was then refitted from scratch. This entire permutation procedure was repeated across 20 independent iterations, and the resulting metrics were averaged.

\subsection{Disease-related community differences}\label{sup:interp-org}

\paragraph{Measures} To quantify the disease-related differences in community organization, we first formalize how frequently ROIs within or between functional systems share a community. For any pair of cortical ROIs $i \neq j$, let $B_{ij} = 1$ if they are assigned to the same first-level community and $0$ otherwise. For two functional systems $a$ and $b$, the raw proportion of ROI pairs sharing a community is the mean $\bar{B}_{ab}$ across all pairs with $i \in a$ and $j \in b$. However, because having fewer or larger communities naturally increases the chance of any two ROIs being grouped together, we correct $\bar{B}_{ab}$ for random co-assignment. Specifically, the expected probability that two cortical ROIs share a community by chance is given by
\begin{equation}
q = \frac{\sum_k n_k(n_k - 1)}{400 \times 399},
\end{equation}
where $n_k$ is the number of cortical ROIs assigned to community $k$. Using $q$ as a baseline, the chance-corrected co-assignment between systems $a$ and $b$ is defined as
\begin{equation}
R_{ab} = \frac{\bar{B}_{ab} - q}{1 - q}.
\end{equation}
Here, $R_{ab} = 0$ indicates that the two systems share communities no more often than expected from community sizes alone, whereas $R_{ab} > 0$ indicates preferential grouping between systems $a$ and $b$, and the diagonal term $R_{aa}$ measures the internal cohesion of system $a$. Consequently, when comparing patients against healthy controls, an increase in $R_{ab}$ ($a \neq b$) captures the blurring of functional boundaries between two distinct systems, whereas a decrease in $R_{aa}$ reflects the fragmentation within a single system. For each participant, we compute $R_{ab}$ for all 28 pairs of the Yeo-7 networks and all 153 pairs of the Yeo-17 networks to evaluate system-level alterations, and we quantify overall alignment between the participant's partition and the canonical Yeo-17 atlas using the NMI defined in Section~\ref{sup:comm}.

\paragraph{Group comparison}
To isolate disease-related changes from confounding factors, we regress each community measure ($R_{ab}$ and NMI) on diagnostic group alongside recording site and the full set of covariates defined in Section~\ref{sup:affil} (evaluated at level $l=1$). Including these covariates in the regression model controls for baseline variations driven by covariates and allows us to evaluate the adjusted difference between patients and healthy controls directly from the $t$ statistic (and corresponding $p$ value) of the diagnostic term.

\paragraph{Discovery and replication}
Because testing all 28 Yeo-7 pairs and 153 Yeo-17 pairs introduces substantial multiple testing, we select candidate differences on one set of sites and test them on another. Specifically, the sites of each cohort are partitioned into a discovery set and a held-out replication set. In the discovery set, we rank the candidate differences within each family of measures by their $|t|$ statistics and carry the leading differences forward with their observed directions of effect. The NMI comparison with the Yeo-17 atlas was planned before the analysis and is carried forward without this selection. In the held-out replication set, each selected difference is tested once using a one-sided $t$ test in its predicted direction (resulting in the replication $p$ values reported in the main text, such as those for control--default mixing and somatomotor cohesion). In ADNI, MCI participants are assigned to the discovery or replication set of their respective recording site. To test whether community disruptions progress monotonically across disease stages (CN $\to$ MCI $\to$ AD), we perform a trend regression across the held-out sites by coding the disease predictor of CN, MCI, and AD as $0$, $0.5$, and $1$, respectively.

\paragraph{Visualization in Figs.~\ref{fig:ad-cn} and ~\ref{fig:consensus}}
To localize the system-level differences shown in Fig.~\ref{fig:ad-cn} to individual cortical regions, we adapt the system-level metric $R_{ab}$ to the ROI level. Specifically, the chance-corrected co-assignment of a cortical ROI $i$ with a partner system $b$ is defined as
\begin{equation}
r_i(b) = \frac{f_i(b) - q}{1 - q},
\end{equation}
where $f_i(b)$ is the proportion of cortical ROIs in system $b$ (excluding $i$ itself) that fall into the same community as ROI $i$. We fit the regression model described above (including MCI and AD indicators and all covariates) to $r_i(b)$ across all 1,732 ADNI participants, and color each ROI in Fig.~\ref{fig:ad-cn} by the resulting $t$ statistic comparing AD against CN.

For the left panel of Fig.~\ref{fig:consensus}, to show representative community maps for CN, MCI, and AD, we draw 50 random samples of 197 participants per group (sampling with replacement for AD). For each sample, we average the first-level co-assignment matrices across participants and cluster the resulting matrix via average linkage into 17 group-level communities, coloring each by its majority Yeo-7 network. We then display the sample whose NMI with the Yeo-17 atlas is closest to the group median. For the right panel of Fig.~\ref{fig:consensus}, we measure atlas alignment for each participant as the average fraction of an ROI's community members that belong to the same Yeo-17 network as the ROI. Differences in this alignment metric for MCI and AD relative to CN are then estimated across all 1,732 ADNI participants using the same regression covariates.

\subsection{Regional evidence}\label{sup:interp-model}

To examine whether IC-GNN's predictions reflect the observed community organization, we analyze the regional evidence $c_i$ defined in Section~\ref{sec:readout}, where a participant's prediction logit decomposes additively as $w_f + \sum_{i} c_i$. For each participant, $c_i$ is averaged across the three trained models (one per random seed) in which that participant belonged to the held-out test set. To summarize evidence at the system level, the 419 ROIs are grouped into eight systems: the seven cortical Yeo networks (with the temporal--parietal network grouped into the default network) and one system comprising all noncortical ROIs.

\paragraph{Network share (Fig.~\ref{fig:evidence})}
To quantify how much each region and functional system contributes to the diagnostic separation in Fig.~\ref{fig:evidence}, we first compute the regional evidence difference $\Delta_i$ between patients and controls for each ROI $i$. On ADNI, $\Delta_i$ is the mean difference in $c_i$ between AD and CN participants; on ABIDE, $\Delta_i$ is the regression coefficient on the ASD indicator after adjusting for the covariates of Section~\ref{sup:interp-org} (with ROIs surviving Benjamini--Hochberg FDR $q < 0.05$ outlined in Fig.~\ref{fig:evidence}). We then normalize each ROI's difference by the whole-brain total to obtain its percentage share, $100\,\Delta_i / \sum_j \Delta_j$, which colors the brain maps in Fig.~\ref{fig:evidence}. In the accompanying bar plots, each bar shows a system's total share of the evidence difference (the sum of shares over its constituent ROIs, such as the 40\% control network share in ADNI and 37\% default network share in ABIDE), while the reference tick marks the baseline share expected purely from the proportion of the 419 ROIs belonging to that system.

\paragraph{Community organization and evidence}
Finally, to test whether IC-GNN tracks continuous individual variations in network blurring across held-out ADNI participants, we regress the model's summed evidence across the 156 control and default ROIs (outcome) on the participant's first-level control--default co-assignment $R_{\text{control, default}}$ (predictor). Both the predictor and the outcome are $z$-scored across participants so that the regression slope yields a standardized coefficient $\beta$. We fit this regression adjusting for the covariates of Section~\ref{sup:interp-org} both without diagnosis (testing the overall association across all held-out participants, $\beta = 0.18$) and with diagnosis included as an additional covariate (testing whether the association holds even among participants within the same diagnostic group, $\beta = 0.13$). Statistical significance is assessed using 9,999 within-site Freedman--Lane permutations.

\end{document}